\documentclass[journal]{IEEEtran}
\usepackage{amsmath,amsfonts}
\usepackage{algorithmic}
\usepackage{algorithm}
\usepackage{array}
\usepackage[caption=false,font=normalsize,labelfont=sf,textfont=sf]{subfig}
\usepackage{textcomp}
\usepackage{stfloats}
\usepackage{url}
\usepackage{verbatim}
\usepackage{graphicx}
\usepackage{cite}
\usepackage{multirow}
\usepackage{booktabs}
\usepackage{makecell}
\usepackage{hyperref}
\hypersetup{colorlinks=true,
            linkcolor=blue,
            anchorcolor=blue,
            citecolor=blue}
\begin{document}

\title{USER: Unified Semantic Enhancement with Momentum Contrast for Image-Text Retrieval}

\author{Yan Zhang, Zhong Ji,~\IEEEmembership{Senior Member,~IEEE,} Di Wang, Yanwei Pang,~\IEEEmembership{Senior Member,~IEEE,} \\
        Xuelong Li,~\IEEEmembership{Fellow,~IEEE}

\thanks{Manuscript received xxx xx, 2023; revised xxx xx, 2023.}
\thanks{This work was supported by the National Natural Science Foundation of
China under Grants 62176178.}
\thanks{Yan Zhang, Zhong Ji, Di Wang*(corresponding author), and Yanwei Pang are with the School of Electrical and
Information Engineering; Tianjin Key Laboratory of
Brain-inspired Intelligence Technology, Tianjin University, Tianjin 300072,
China (e-mail:
yzhang1995@tju.edu.cn; jizhong@tju.edu.cn; wangdi2015@tju.edu.cn; pyw@tju.edu.cn).}
\thanks{
  Xuelong Li is with School of Artificial Intelligence, OPtics and ElectroNics
(iOPEN) and the Key Laboratory of Intelligent Interaction and Applications, Ministry of Industry and Information Technology, 
Northwestern Polytechnical University, Xi'an 710072, P.R. China 
(e-mail: li@nwpu.edu.cn).}
}

\markboth{Journal of \LaTeX\ Class Files,~Vol.~14, No.~8, August~2021}%
{Shell \MakeLowercase{\textit{et al.}}: A Sample Article Using IEEEtran.cls for IEEE Journals}


\maketitle

\begin{abstract}
  As a fundamental and challenging task in bridging language and vision domains, Image-Text Retrieval (ITR) aims at searching for the target instances 
  that are semantically relevant to the given query from the other modality, and its key challenge is to measure the semantic similarity across different modalities. 
  Although significant progress has been achieved, existing approaches typically suffer from two major limitations: (1) It hurts the accuracy of the representation by
  directly exploiting the bottom-up attention based region-level 
  features where each region is equally treated. (2) It limits the scale of negative sample pairs by employing the mini-batch based end-to-end training mechanism. 
  To address these limitations, we propose a Unified Semantic Enhancement Momentum Contrastive Learning (USER) method for ITR. Specifically, we delicately design two 
  simple but effective Global representation based Semantic Enhancement (GSE) modules. 
  One learns the global representation via the self-attention algorithm, noted as Self-Guided Enhancement (SGE) module. 
  The other module benefits from the pre-trained CLIP module, which provides a novel scheme to exploit and transfer the knowledge from an off-the-shelf model, 
  noted as CLIP-Guided Enhancement (CGE) module. 
  Moreover, we incorporate the training mechanism of MoCo into ITR, in which two dynamic queues are employed to enrich and enlarge the scale of negative sample pairs. 
  Meanwhile, a Unified Training Objective (UTO) is developed to learn from mini-batch based and dynamic queue based samples. 
  Extensive experiments on the benchmark MSCOCO and Flickr30K datasets demonstrate the superiority of both retrieval accuracy and inference efficiency. 
  For instance, compared with the existing best method NAAF, the metric R@1 of our USER on the MSCOCO 5K Testing set is improved by 5\% and 2.4\% on caption retrieval and image 
  retrieval without any external knowledge or pre-trained model while enjoying over 60 times faster inference speed. Our source code will be released at \url{https://github.com/zhangy0822/USER}.
\end{abstract}


\section{Introduction}
\begin{figure}[t]
  \centering
  \includegraphics[width=0.95\columnwidth]{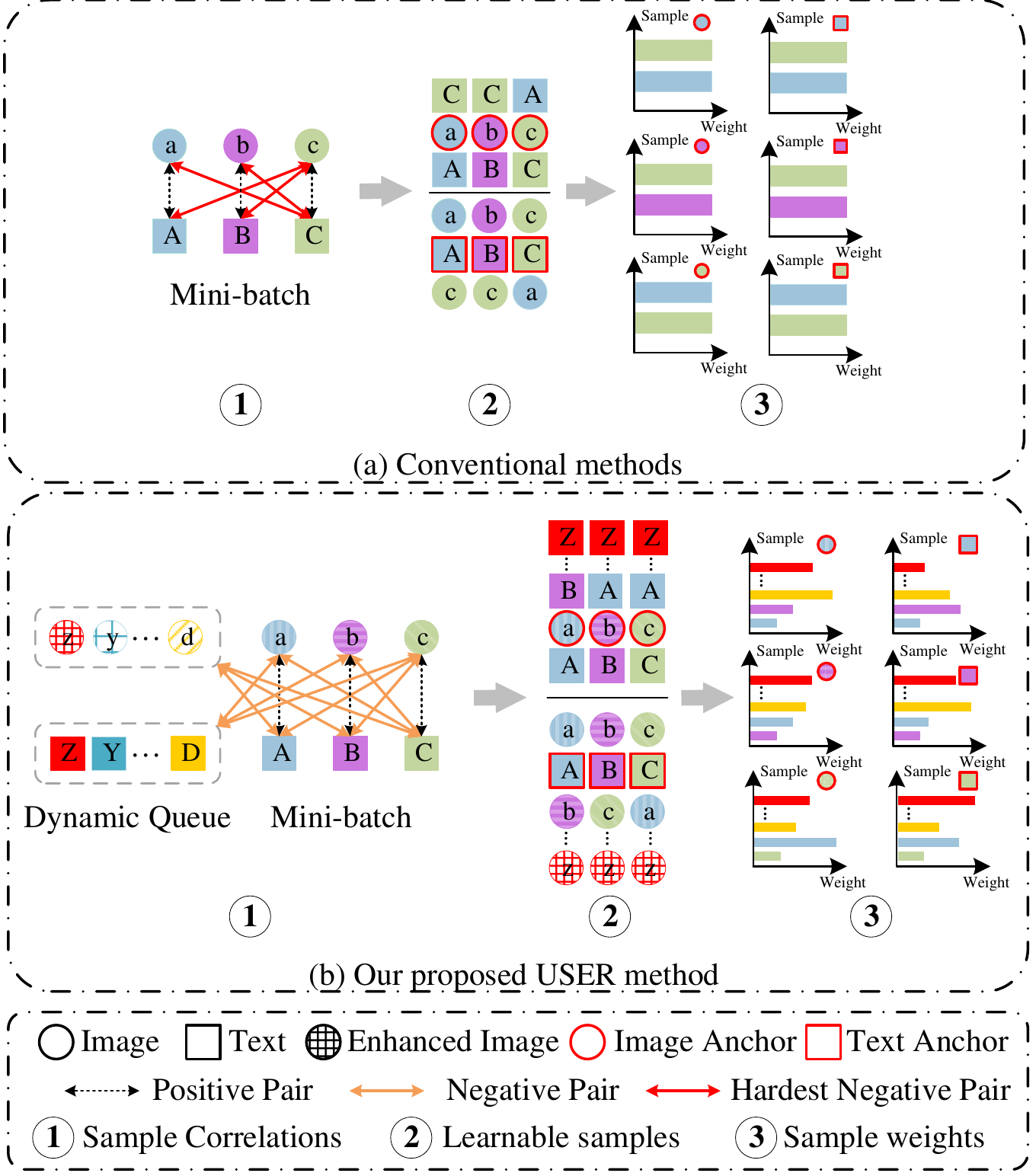} 
  \caption{Comparison of our USER method with the conventional methods, where the three stages: the correlations among samples, the learnable samples for each anchor sample, and
  the sample weights for training are represented. Note that the identical colors represent the matched sample pairs, 
  and the lowercase letters and circles represent images while the uppercase letters and squares represent texts.
  Particularly, (a) Existing methods train the model with conventional training paradigm, such as \cite{lee2018stacked,li2019visual,wang2020consensus,jistep,diao2021similarity,chen2021learning,zhang2022negative},
  which treat the bottom-up attention \cite{anderson2018bottom} based region features equally  
  and only regard the hardest negative sample in a mini-batch as negative sample with the same weights. 
  Differently in (b), the proposed USER employs the enhanced visual features, which are guided by itself or a pretrained CLIP \cite{radford2021learning} module, 
  and introduces two dynamic queues on-the-fly to enlarge the scale of negative samples with the different weights.}
  \label{fig1}
\end{figure}

\IEEEPARstart{I}{mage}-Text Retrieval (ITR) aims at searching for semantically relevant images from an image database for a given text query and vice versa. 
Nowadays, ITR has attracted extensive attention \cite{faghri2018vse++,lee2018stacked,li2019visual,wang2020consensus,jistep,chen2021learning,diao2021similarity, zhang2022negative} since there is an excellent benefit to 
extensive relevant applications, such as search engines and multimedia data management systems.
The critical challenges of ITR lie in accurately obtaining visual and textual semantic representations, and learning a optimal joint embedding space from adequate sample pairs.
Despite tremendous efforts have been dedicated to tackling the above challenges, such as the global-level matching methods \cite{faghri2018vse++,li2019visual,chen2021learning,li2022image},
attention-based local-level matching methods \cite{lee2018stacked,wang2022coren,chen2020imram,jistep,dong2022hierarchical}, external knowledge information based methods \cite{wang2020consensus,wang2022coder,shi2019knowledge,ge2022cross},
it remains challenging due to the inaccuracy of semantic representation and limited negative image-text pairs in a mini-batch.

In the past decade, we have witnessed how the paradigm shifts from hand-crafted methods \cite{pereira2013role, wang2015joint} to deep neural networks methods 
\cite{lee2018stacked,li2019visual}, where the latter significantly improves the retrieval performance since its powerful representation and reasoning ability.
Actually, extracting the bounding box based region-level features from the bottom-up attention \cite{anderson2018bottom} has become standard practice for most cross-modal tasks, 
such as visual question answering (VQA) \cite{anderson2018bottom}, ITR \cite{lee2018stacked, wang2020consensus}, image captioning \cite{luo2021dual}, and recent popular vision-language pre-training works \cite{lu2019vilbert,li2020oscar,chen2020uniter}.
As a pioneering work in ITR, SCAN \cite{lee2018stacked} employs the advantage of bottom-up attention \cite{anderson2018bottom},
represents each image as a set of region-level features, and attends to relevant fragments from another modality to discover all word-region alignments.
Afterwards, various variants based on SCAN \cite{lee2018stacked} are proposed to boost the retrieval performance by cross-modal interaction \cite{jistep, ji2019saliency, wang2022coren,zhang2022latent},
position attention \cite{wang2019position}, and graph neural networks \cite{liu2020graph,peng2022relation,ji2022heterogeneous,diao2021similarity}. 
Aside from these attention-based local-level matching methods, there are several research lines that also directly employ region-level features to achieve cross-modal alignment
by holistic embeddings \cite{faghri2018vse++, li2019visual, chen2021learning,wang2020consensus,wang2022coder} or knowledge-aided representation learning \cite{wang2020consensus,wang2022coder,shi2019knowledge}. 
However, they neglect the impact of global semantic information when encoding region-level visual features since some objects are not precisely identified by the object detector or even absent \cite{cao2022vision}.

In addition, the current end-to-end training mechanism limits the scale of negative sample pairs in a mini-batch, as shown in Fig. \ref{fig1}(a). Generally speaking, the batch size in ITR is 128 or 256 for 
achieving good retrieval performance. More importantly, the widely employed bi-directional triplet ranking loss \cite{faghri2018vse++} only simply considers the hardest negative 
sample within a mini-batch, and employs the same weights for positive and negative pairs, which significantly degrades the role of other samples. 
On the contrary, Chen \textit{et al.} \cite{chen2020adaptive} argued that each sample has different importance for learning joint embedding space.
The InfoNCE \cite{oord2018representation} loss has been widely adopted \cite{chen2020simple, liu2021hit, han2021text, radford2021learning},
where all the positive and negative pairs in a mini-batch are leveraged to learn representations. To mitigate the hubness problem in the high-dimensional embedding space, 
Liu \textit{et al.} \cite{liu2020hal} proposed the hubness-aware loss function, which also considers all samples.  
Actually, a set of self-supervised methods \cite{chen2020simple,he2020momentum,chen2020improved,grill2020bootstrap} have emphasized the necessity of large-scale negative samples in the contrastive learning field. 
Meanwhile, there are already various advanced approaches for several downstream tasks \cite{wang2022coder,liu2021hit,han2021text}.

To this end, we propose a Unified Semantic Enhancement Momentum Contrastive Learning (USER) method for ITR by incorporating the enhancement of visual region-level features with 
Momentum Contrast \cite{he2020momentum}. Specifically, we propose two alternative Global representation based Semantic Enhancement (GSE) modules to 
enhance the region-level features, where the Self-Guided Enhancement (SGE) module employs the self-attention algorithm \cite{vaswani2017attention} and the CLIP-Guided Enhancement (CGE) module 
provides a novel scheme to employ and transfer the knowledge from the pre-trained cross-modal CLIP \cite{radford2021learning} model. 
Moreover, we introduce the MoCo \cite{he2020momentum} paradigm into the training process of ITR to enlarge the scale of negative sample pairs on-the-fly, as shown in Fig. \ref{fig1}(b).
The superiority of our proposed USER is verified on the two popular benchmarks, i.e., Flickr30K\cite{young2014image} dataset 
and MSCOCO \cite{lin2014microsoft} dataset.

Our main contributions are summarized as follows:
\begin{itemize}
  \item Instead of equally treating region-level features that are extracted from the bottom-up attention, we propose two Global representation based Semantic Enhancement (GSE) modules 
  for ITR by exploiting the self-attention algorithm and the pre-trained CLIP model, respectively.
  \item Based on the proposed GSE module, a novel training paradigm with Momentum Contrastive Learning is efficiently introduced to enlarge the scale of negative sample pairs.
  \item We further incorporate and improve the current hubness-aware loss function with Momentum Contrastive Learning as a Unified Training Objective (UTO), 
  where two dynamic queues and two key encoders for two modalities are introduced to maintain representation consistency.   
\end{itemize}

The remaining part of this paper is organized as follows. We present the related works about USER in Section II. The main
components: Image Encoders, Text Encoders, Momentum Cross-modal Contrast Learning, followed by Unified Training Objective, are introduced in Section III.
Extensive experiments conducted on the MSCOCO and Flickr30K datasets are shown in Section IV. Finally, we conclude our USER method in Section V.

\section{Related work} 
\subsection{Image-Text Retrieval}
Image-Text Retrieval, which aims at searching for instances from another modality as query, has dramatically revolutionized over the last few years with the renaissance of deep learning.
According to the calculation method of similarity, the current methods can be roughly categorized into two groups, i.e., global-level matching methods and local-level matching methods.

\textbf{Global-level matching methods} represent the image and text as holistic embeddings and perform dot products between them. Enabled by the CNN and Skip-Gram architecture, early methods 
employ Canonical Correlation Analysis (CCA) \cite{rasiwasia2010new} and its follow-up variants \cite{wang2016multimodal,xu2017learning} to represent image or text as a global representation.
Kiros \textit{et al.} \cite{kiros2014unifying} employed the bidirectional triplet ranking loss to optimize the projection matrices. Faghri \textit{et al.} \cite{faghri2018vse++}
only paid attention to the hardest negative sample to improve the performance. Afterwards, Li \textit{et al.} \cite{li2019visual, li2022image} performed 
region and word relationship reasoning via Graph Convolutional Networks \cite{kipf2017semi} (GCN) and Gated Recurrent Units \cite{bahdanau2015neural} (GRU). Wang \textit{et al.} \cite{wang2020consensus} 
extracted consensus information from large-scale external knowledge and employed GCNs to learn the aided consensus-aware concept representation. Chen \textit{et al.} \cite{chen2021learning} proposed a plug-and-play
Generalized Pooling Operator (GPO), which automatically aggregates itself into holistic embeddings for different features. 
Zhang \textit{et al.} \cite{zhang2020deep} developed a two branches architecture for image encoding, where the global embedding is embedded by an additional ResNet-152 \cite{he2016deep} model.  

Our USER method also falls in this group. Differently, in USER, the region-level features are enhanced by a self-attention \cite{vaswani2017attention} algorithm 
or an alternative pre-trained model \cite{radford2021learning} based method. Moreover, we train the optimal joint embedding space by applying two dynamic queues to
learn from adequate negative sample pairs.

\textbf{Local-level matching methods} represent the image and text as two sets of local-level embeddings,
typically region-level features for images and word-level features for texts. The overall similarity is inferred via the fine-grained alignment among these local-level features.
Flagship method SCAN \cite{lee2018stacked} first employs the bottom-up attention \cite{anderson2018bottom} to detect the salient object and introduce the stacked cross attention algorithm to obtain the similarity,
which motivates various sophisticated methods \cite{jistep, wang2019position, chen2020imram, wen2020learning, wang2022coren}.
Another branch of work focuses on employing the GCNs to obtain more precise representation \cite{liu2020graph,jing2021learning} or reason more accurate similarity \cite{diao2021similarity,peng2022relation}.
Unlike the above unilateral attention mechanism, Zhang \textit{et al.} \cite{zhang2022negative} proposed a negative-aware attention framework, which calculates the dissimilarity degree of mismatched fragments.  
Although these sophisticated and well-targeted models have made great progress, the prohibitive inference speed limits their real-world usage, which is seldom investigated but crucial, especially 
in real application scenarios.

\subsection{Semantic Enhancement Methods for ITR}
The dominant task of ITR is learning a semantic embedding space where paired samples are closer than unpaired samples. More importantly, we need to enhance the visual semantics to learn precise representation, whereas textual modality does not. 
The reason may stem from the difference in their respective signal spaces. Particularly, the visual modality
has continuous and high-dimensional spaces, while the textual
modality has discrete signal spaces (i.e., words, sub-word units) for building tokenized dictionaries.  
Song \textit{et al.} \cite{song2019polysemous} argued that the semantic ambiguity exists in the visual modality. 
Recently, semantic enhancement based methods have been dramatically developed, such as 
memory based methods \cite{li2021memorize,ji2022heterogeneous} and external tools based methods \cite{wang2022coder, wang2020consensus,ge2022cross}.
For example, Li \textit{et al.} \cite{li2021memorize} introduced global memory banks to store features across two modalities, which are employed to enhance the feature representations.
Ji \textit{et al.} \cite{ji2022heterogeneous} proposed a heterogeneous memory enhanced graph reasoning network to learn more discriminative and robust representations.
Ge \textit{et al.} \cite{ge2022cross} designed the intra-modal spatial and semantic graphs to enhance their semantic representations, where the visual scene graphs are generated by the off-the-shelf Neural Motifs \cite{zellers2018neural} tool.
The existing Transformer-based methods almost design a type of cross-modal interaction architecture to complement or enhance the modality-specific representation, such as 
\cite{wei2020multi,dong2022hierarchical,wang2022coren}. 
In this paper, we develop two alternative semantic enhancement methods to enhance the semantic representation. One employs the superiority of the self-attention algorithm \cite{vaswani2017attention} 
to enhance the visual representation without any cross-modal operation. 
Unlike the above complicated and well-targeted methods, the other method incorporates the pre-trained CLIP \cite{radford2021learning} model into ITR, 
which studies how to transfer and employ the knowledge from the pre-trained model. 

\subsection{Contrastive Learning Methods}
Contrastive Learning aims at mapping the embeddings of positive pairs together while pushing those of negative pairs
apart, which has made excellent progress in unsupervised visual representation learning \cite{chen2020simple,he2020momentum,chen2020improved,grill2020bootstrap}
and vision-language pre-training \cite{radford2021learning,kim2021vilt,li2021align,wang2021vlmo}.
He \textit{et al.} \cite{he2020momentum} proposed a novel Momentum Contrastive (MoCo) training paradigm for facilitating contrastive unsupervised learning, which employs a momentum-updated key encoder and a dynamic dictionary to build a large and consistent dictionary on-the-fly.
Afterwards, many studies have attempted to introduce MoCo in their downstream tasks. For instance, Liu \textit{et al.} \cite{liu2021hit} proposed a Hierarchical Transformer model for video-text retrieval task, which performs
cross-modal contrastive matching and employs the InfoNCE loss \cite{oord2018representation} as its training objective. Xiao \textit{et al.} \cite{han2021text} developed a text-based person search method,
which achieves identity-level contrastive learning. Wang \textit{et al.} \cite{wang2022coder} invented a diversity-sensitive contrastive learning loss for instance-level and concept-level representations,
where the latter is extracted from external knowledge. 
Unlike the above methods, we incorporate MoCo and a unified training objective into ITR for training our semantic enhancement modules without leveraging any external knowledge,
where two types of samples contribute to obtaining precise modality-specific representations. 

\section{Methods}
\subsection{Overall Architecture}
\begin{figure*}[!htb] 
  \centering
  \includegraphics[width=0.95\textwidth]{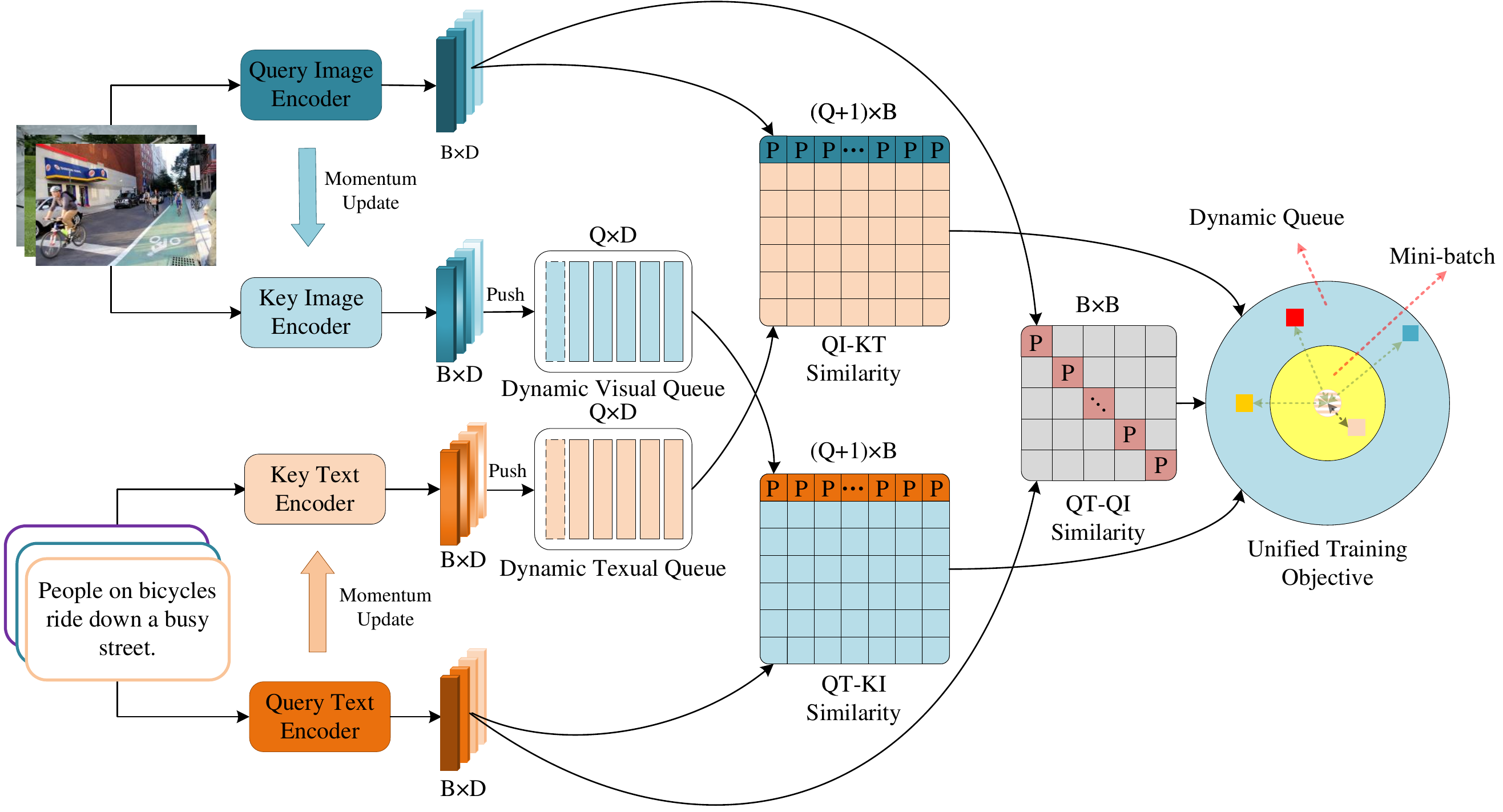} 
  \caption{Illustration of the proposed USER method. The Key Encoders are momentum-updated, while the Query Encoders are gradient-updated. The Dynamic Queues (DQ) are built to
  store the key embeddings of two modalities from the Key Encoders with the latest mini-batch enqueued and the earliest mini-batch dequeued. Symbol ``P'' means the positive sample pairs.}
  \label{fig2}
  \end{figure*}

In this section, we elaborate on our Unified Semantic Enhancement Momentum Contrastive  Learning (USER) method for ITR, as illustrated in Fig. \ref{fig2}.
It consists of two query encoders $f_q^I$, $f_q^T$ (Query Image Encoder and Query Text Encoder) along with two key encoders $f_k^I$, $f_k^T$ (Key Image Encoder and Key Text Encoder),
parameterized by $\theta_q^I$, $\theta_q^T$, $\theta_k^I$, $\theta_k^T$, respectively. 
Note that the structures of the query encoders and the key encoders are identical for the two modalities. 
Additionally, we leverage two Dynamic Queues to store more visual and textual embeddings,
which will be discussed later. During training, both the query encoders and 
the key encoders are employed to generate the corresponding embeddings for their modalities, and the output of the key encoders is pushed into their corresponding dynamic queues,
which aims at enlarging the scale of negative pairs for contrastive learning.
During inference, only the query encoders are employed to generate the corresponding embeddings for the similarity computing.

In the following subsections, we first introduce the Image Encoders and Text Encoders in Section III-B and Section III-C, including our proposed two alternative 
Global representation based Semantic Enhancement (GSE) modules.
Afterwards, we delineate the Momentum Cross-modal Contrast Learning in Section III-D.
Finally, the Unified Training Objective (UTO) is presented in Section III-E.

\subsection{Image Encoders}
As shown in Fig. \ref{fig3}, the image encoder consists of three main components: Feature Extraction, Global representation based Semantic Enhancement (GSE), and Feature Aggregation.

\subsubsection{\textbf{Feature Extraction}}
Given an image $I$, we employ the bottom-up attention \cite{anderson2018bottom} to present it as a set of region-level features $\textbf{V}=\left\{v_i|i\in[1,K],v_i\in \mathbb{R}^{d^I}\right\}$,
where $d^I$ and $K$ denote the dimension of visual features and the number of regions, respectively. 

\subsubsection{\textbf{Global representation based Semantic Enhancement}}
Most existing approaches model the relationships among object regions and perform local-level or global-level matching directly \cite{lee2018stacked,li2019visual,diao2021similarity}.
However, they neglect the fact that some objects are not precisely identified or even absent. 
To encode region-level visual features more accurately, we propose to employ global representation to enhance the region-level features.
To this end, we shift our attention to learn the global semantic representation that equips discriminative global information. Concretely, two alternative modules are proposed, 
which are noted as Self-Guided Enhancement (SGE) module and CLIP-Guided Enhancement (CGE) module.

\begin{figure*}[!htb]
  \centering
  \includegraphics[width=0.95\textwidth]{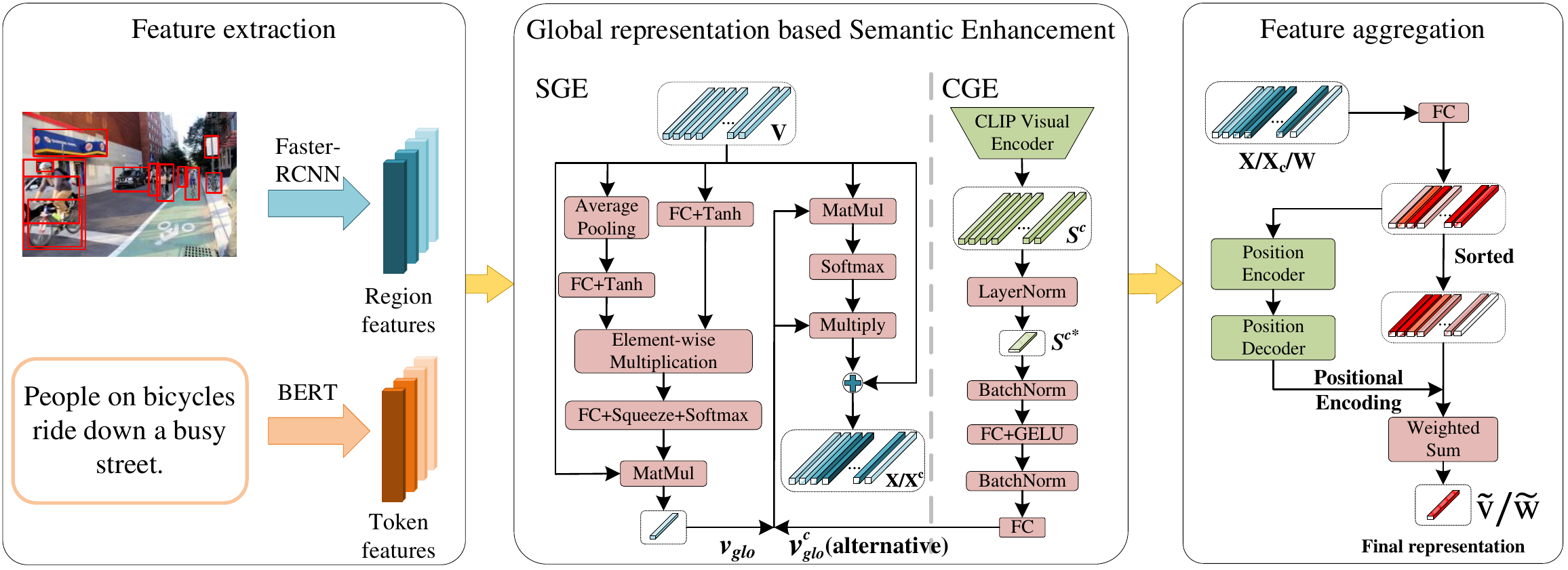} 
  \caption{Illustration of our Image Encoders and Text Encoders. The left box shows the extraction of two modality features. The middle box shows our proposed
  GSE module, which contains two methods: SGE and CGE. Note that the SGE and CGE are not employed together in the entire framework. The right box shows how to aggregate the local-level features
  into holistic embeddings. }
  \label{fig3}
\end{figure*}

(a) Self-Guided Enhancement (SGE) module. Given the region-level features \textbf{V}, we perform the self-attention algorithm \cite{vaswani2017attention} over local image regions 
to obtain global visual representation, as shown in Fig. \ref{fig3}.
Specifically, we first leverage the average feature as an initial global embedding $v_{ave}$, which is computed as follows:
\begin{equation}
  v_{ave} = \frac{1}{K}\sum_{i=1}^K v_i. \label{eq1}  
\end{equation}

Then, two mapping functions $\kappa (\cdot)$ and $\nu (\cdot)$, termed region-level features mapping function and 
global embedding mapping function, are employed: 
\begin{equation}
  \kappa (v_i) = \textbf{W}_\kappa v_i + b_\kappa, \nu (v_{ave}) = \textbf{W}_\nu v_{ave} + b_\nu,  \label{eq20}
\end{equation}
where $\textbf{W}_\kappa \in \mathbb{R}^{d^I \times d^I}$ and $\textbf{W}_\nu \in \mathbb{R}^{d^I \times d^I}$ represent the learnable parameters matrices, 
$b_\kappa$ and $b_\nu$ represent the bias terms.
All regions are aggregated with Element-wise
Multiplication and Softmax function, which are defined by the following equations:
\begin{equation}
  r_{i} = TB(\nu (v_{ave})) \odot TB(\kappa (v_i)), \label{eq2}
\end{equation}
\begin{equation}
  s_{i} = \frac{exp(\textbf{W}_a r_{i})}{\sum_{i=1}^K exp(\textbf{W}_a r_{i})}, \label{eq3}
\end{equation}
\begin{equation}
  v_{glo} = \left \| \sum_{i=1}^K s_j v_i \right \|_2, \label{eq4}
\end{equation}  
where $TB$ denotes the combination of Tanh activation function and Batch Normalization, $\odot$ denotes the Element-wise Multiplication, 
and $\textbf{W}_a\in \mathbb{R}^{{d^I} \times 1}$ is a learnable parameter matrix.
The weight coefficient $s_i$ represents the relevance between $v_i$ and $v_{ave}$.

After obtaining the global embedding, an attention mechanism is performed to enhance the region-level features:
\begin{equation}
  a_i = \frac{exp(v_{glo}^T v_i)}{\sum_{i=0}^K exp(v_{glo}^T v_i)}, \label{eq5}
\end{equation}
\begin{equation}
  x_i = v_i + a_i^T\cdot v_{glo}, \label{eq6}
\end{equation}
where $a_i$ represents the importance score between region $i$ and the global representation $v_{ave}$.
Finally, $\textbf{X}=\{x_i|i \in [1,K], x_i \in \mathbb{R} ^{d^I}\}$ represents the SGE based region-level features.

(b) CLIP-Guided Enhancement (CGE) module. Generally speaking, we extract features by employing image and text encoders, which are pre-trained on unimodal data, 
typically the ResNet101 \cite{he2016deep} pre-trained on ImageNet \cite{deng2009imagenet} and the BERT \cite{kenton2019bert} pre-trained on large corpora. 
However, global visual semantic information is often difficult to present owing to the limited data compared with WIT \cite{radford2021learning}, even though most methods apply complex
network structures, such as GCNs \cite{wang2020consensus,diao2021similarity,li2022image} and Transformer \cite{wei2020multi,li2021memorize,wang2022coren,dong2022hierarchical}. 
To mitigate this flaw and validate the importance of the discriminative global semantic representation, we employ a transfer learning strategy for extracting the global visual embedding.

Concretely, as shown in Fig. \ref{fig3}, we first apply the CLIP Image Encoder \cite{radford2021learning} to extract the spatial visual feature, 
which is reshaped to $s^c\in 
\mathbb{R}^{HW \times d^I_c}$, where $d^I_c$ denotes the dimensionality of spatial visual features, and $HW$ denotes the number of channels.
Then, we employ Layer Normalization to transform $s^c \in \mathbb{R}^{HW \times d^I_c}$ to $s^{c*} \in \mathbb{R}^{1 \times d^I_c}$. 
Afterwards, two fully-connected layers with Batch Normalization function are defined to transform $s^{c*} \in \mathbb{R} ^{d_c^I}$ to $v^c_{glo} \in \mathbb{R} ^{d^I}$ as follows:
\begin{equation}
  v^c_{glo} = \textbf{W}_{c2}BN(\sigma((\textbf{W}_{c1}BN(s^{c*})+b_{c1}))) + b_{c2}, \label{eq7}
\end{equation}
where $\textbf{W}_{c1} \in \mathbb{R}^{d^I_c \times d^I}$ and $\textbf{W}_{c2}\in \mathbb{R}^{d^I \times d^I}$ represent the learnable parameters matrices, 
$b_{c1}$ and $b_{c2}$ represent the bias terms, and $BN$ and $\sigma$ denote the Batch Normalization and GELU activation function, respectively.

Then, we denote $v^c_{glo}$ as the CLIP-based global embedding, and additionally apply Eq.\eqref{eq5} and Eq.\eqref{eq6} to enhance the region-level 
features as the SGE module, denoted as $\textbf{X}^c=\{x_i^c|i\in [1,K],x_i^c\in\mathbb{R} ^{d^I}\}$.

\subsubsection{\textbf{Feature Aggregation}}

We employ a fully connected layer to transform $\textbf{X}$ ($\textbf{X}^c$) to joint embedding space $\hat{\textbf{V}}$ ($\hat{\textbf{V}}^c$):
\begin{equation}
  \hat{v}_i = \textbf{W}_{fv} x_i + b_{fv}, \label{eq8}
\end{equation}
where $\textbf{W}_{fv} \in \mathbb{R}^{d^I \times d^J}$, $b_{fv}$ are the learnable parameters matrix and the bias term, respectively, and $d^J$ is the dimensionality of the
joint embedding space.

After obtaining the enhanced region-level features, we aggregate these features to be a holistic embedding by following the operator \cite{chen2021learning},
which is defined as:
\begin{equation}
  \widetilde{v}=\sum_{k=1}^N \theta_k \cdot max_k(\{\hat{v}_i\}_{i=1}^N), where \sum_{k=1}^K \theta_k = 1, \label{eq9}
\end{equation}
where $N$ denotes the number of feature vectors, and it refers the number of regions $K$ in visual encoding, $\theta_k$ represents the coefficient for the $k$-th maximum value among the $N$ 
elements. To this end, we are committed to learning a parameterized coefficient generator, which consists of two major components: (a) Position Encoder. 
A positional encoding function based on trigonometric function, (b) Position Decoder. A pooling coefficients generation function based on BiGRU \cite{bahdanau2015neural}. 
The architecture is shown in Fig. \ref{fig3}. 
 
\textbf{Position Encoder.} We follow \cite{chen2021learning,wang2022coder,vaswani2017attention} and employ the trigonometric function to encode each position index $t$ to a dense vector $p_t$, which is leveraged in Transformer \cite{vaswani2017attention} 
and defined as follows:
\begin{equation}
  p_t^i=\begin{cases}
    \sin(w_j,t), \quad when \ i=2j, \\
    \cos(w_j,t), \quad when \ i=2j+1, 
  \end{cases}
  \forall i, \label{eq10}
\end{equation}
where $w_j=\frac{1}{10000^{2j/d_t}}$ and $d_t$ is the size of the dimensionality for the positional encoding.

\textbf{Position Decoder.} We feed the obtained dense vector $p_t$ into a Position Decoder, which outputs
the sequence of pooling coefficients $\theta=\{\theta_t\}_{t=1}^N$. Specifically, the Position Decoder consists of a BiGRU and a multi-layer perceptron (MLP):
\begin{equation}
  \{h_t\}_{t=1}^N=BiGRU(\{p_t\}_{t=1}^N),  \label{eq11}
\end{equation}
\begin{equation}
  \theta_k=MLP(h_t), \label{eq21}
\end{equation}
where $h_t$ is the output of the BiGRU at the position $t$. Finally, we aggregate the sorted region-level features $\{\hat{v}_t^{s}\}_{t=1}^N$ into a
holistic visual representation $\widetilde{v}$:
\begin{equation}
  \widetilde{v}=\sum_{t=1}^N \theta_t \cdot \hat{v}_t^s. \label{eq12} 
\end{equation}

\subsection{Text Encoders}
Since a series of words or sub-word units make up the whole sentence without any information missing, we only perform feature extraction and aggregation without semantic enhancement for textual modality.
Pre-trained BERT \cite{kenton2019bert} is adopted as our Text Encoder backbone. Given a text $L$ with $l$ words, we additionally employ a fully-connected layer to transform the extracted word-level features 
$\textbf{W}=\left\{w_i|i\in[1,l],w_i\in \mathbb{R}^{d^T}\right\}$ to $d^J$-dimensional features:
\begin{equation}
  \hat{w}_i = \textbf{W}_{ft} w_i + b_{ft}, \label{eq13}
\end{equation} 
where $\textbf{W}_{ft} \in \mathbb{R}^{d^T \times d^J}$, $b_{ft}$ are the learnable parameters matrix and bias term, respectively.
Then, we employ Eq.\eqref{eq10}-Eq.\eqref{eq12} to aggregate the word-level features to a holistic textual representation $\widetilde{w}$.

\subsection{Momentum Cross-modal Contrast Learning}
One of the limitations of most current ITR methods is that the end-to-end mini-batch based training paradigm limits the negative sample size.
Inspired by MoCo \cite{he2020momentum} and methods of \cite{liu2021hit,han2021text, wang2022coder}, we first build two dynamic queues for saving more the negative sample embeddings dynamically.
\subsubsection{Dynamic Visual Queue} We build a Dynamic Visual Queue (DVQ) for storing the key visual embeddings, which are output by the Key Image Encoder. 
Specifically, in each training iteration, the visual embeddings of 
the current mini-batch are encoded by the Key Visual Encoder $f_k^V$, which are pushed into the DVQ, and 
the earliest mini-batch in DVQ will be dequeued. 
\subsubsection{Dynamic Textual Queue} Similarly, the textual embeddings of each training iteration are pushed into the Dynamic Textual Queue (DTQ), which are 
output by the Key Text Encoder.

Following MoCo \cite{he2020momentum}, the parameters $\theta_k^I$ and $\theta_k^T$ of the two key encoders are momentum-updated:
\begin{equation}
  \begin{aligned}
    \theta_k^V=m\cdot\theta_q^I+(1-m)\cdot\theta_q^I, \\
    \theta_k^T=m\cdot\theta_q^T+(1-m)\cdot\theta_q^T, \label{eq14}
  \end{aligned}
\end{equation}
where $m \in [0,1)$ denotes momentum coefficient, generally taking a large value to maintain the representation consistency in DVQ and DTQ.
Note that the parameters $\theta_q^I$ and $\theta_q^T$ of the two query encoders are updated by back-propagation, which will be discussed in Section III-E.

\subsection{Unified Training Objective}
Aiming at enlarging the scale of negative pairs and make full use of each sample, we improve the existing hubness-aware loss function \cite{liu2020hal} to adapt both mini-batch based and dynamic queue based 
training mechanism, denoted as Unified Training Objective (UTO).
The visual and textual embeddings in a mini-batch are denoted as $\widetilde{\textbf{V}}=\{\widetilde{v}_i|i \in [1,B], \widetilde{v}_i \in \mathbb{R} ^{1 \times d^J}\}$
and $\widetilde{\textbf{W}}=\{\widetilde{w}_i|i \in [1,B], \widetilde{w}_i \in \mathbb{R} ^{1 \times d^J}\}$, respectively, where $B$ is the number of samples within the mini-batch. 
The visual and textual embeddings in DVQ and DTQ are denoted as $\widetilde{\textbf{V}}^Q=\{\widetilde{v}^q_{q_v}|{q_v} \in [1,Q], \widetilde{v}_{q_v}^q \in \mathbb{R} ^{1 \times d^J}\}$
and $\widetilde{\textbf{W}}^Q=\{\widetilde{w}^q_{q_w}|{q_w} \in [1,Q], \widetilde{w}^q_{q_w} \in \mathbb{R} ^{1 \times d^J}\}$, respectively, where $Q$ is the number of samples within the 
dynamic queue. Additionally, we employ $cosine(\cdot,\cdot)$ to compute the similarity between two sample embeddings.

For mini-batch based training, the loss objective $\mathcal{L}_{mini-hal}$ is employed in each mini-batch. 
Specifically, $S_{ii}=cosine(\widetilde{v}_i,\widetilde{w}_i), i \in [1,B]$ denotes the positive pairs,
$S_{mi}=cosine(\widetilde{v}_m,\widetilde{w}_i), m \in [1,i) \cap (i,B] $ and $S_{ni}=cosine(\widetilde{v}_n,\widetilde{w}_i), n \in [1,i) \cap (i,B]$
denote the negative pairs. The $\mathcal{L}_{mini-hal}$ is defined as follows:  
\begin{equation}
  \begin{aligned}
    \mathcal{L}_{mini-hal}=\frac{1}{B} \sum_{i=1}^B(\frac{1}{\gamma}log(1+\sum_{m\neq i}^Be^{\gamma(S_{mi}-\epsilon)})\\
                      +\frac{1}{\gamma}log(1+\sum_{n\neq i}^Be^{\gamma(S_{ni}-\epsilon)})\\
                      -log(1+S_{ii})), \label{eq15}
  \end{aligned}
\end{equation}
where $\gamma$ and $\epsilon$ denote temperature scale and margin, respectively.

For dynamic queue based training, the loss objective $\mathcal{L}_{DQ-hal}$ is calculated by the sum of losses in DVQ and DTQ, which are denoted as $\mathcal{L}_{DVQ-hal}$ and $\mathcal{L}_{DTQ-hal}$.
Specifically, $S_{i^ki}=cosine(\widetilde{v}_i^k,\widetilde{w}_i), i \in [1,B]$ and $S_{ii^k}=cosine(\widetilde{v}_i,\widetilde{w}_i^k), i \in [1,B]$ denote the positive pairs.
Note that $\widetilde{v}_i^k$ and $\widetilde{w}_i^k$ represent the embeddings output from the key encoders $f_k^V$ and $f_k^T$, respectively.
$S_{i{q_v}}=cosine(\widetilde{v}_{q_v}, \widetilde{w}_i)$, $q_v \in [1,Q]$ and $S_{i{q_w}}=cosine(\widetilde{v}_i, \widetilde{w}_{q_w})$, $q_w \in [1,Q]$ denote the negative pairs.
The $\mathcal{L}_{DVQ-hal}$, $\mathcal{L}_{DTQ-hal}$, and $\mathcal{L}_{DQ-hal}$ are defined as follows:
\begin{equation}
  \begin{aligned}
    \mathcal{L}_{DVQ-hal}=\frac{1}{B} \sum_{i=1}^B(\frac{1}{\gamma}log(1+\sum_{{q_v}=1}^Qe^{\gamma(S_{i{q_v}}-\epsilon)})\\
                      -log(1+S_{i^ki})), \label{eq16}
  \end{aligned}
\end{equation}
\begin{equation}
  \begin{aligned}
    \mathcal{L}_{DTQ-hal}=\frac{1}{B} \sum_{i=1}^B(
                      \frac{1}{\gamma}log(1+\sum_{{q_w}=1}^Qe^{\gamma(S_{i{q_w}}-\epsilon)})\\
                      -log(1+S_{ii^k})), \label{eq17}
  \end{aligned}
\end{equation}
\begin{equation}
    \mathcal{L}_{DQ-hal}=\mathcal{L}_{DVQ-hal} + \mathcal{L}_{DTQ-hal}. \label{eq18}
\end{equation}
The final loss is represented as:
\begin{equation}
  \mathcal{L}=\lambda \mathcal{L}_{mini-hal} + \mathcal{L}_{DQ-hal}, \label{eq19}
\end{equation}
where $\lambda$ balances the weight of different loss functions.

\section{Experiments}
 
\subsection{Datasets amd Evaluation Metrics}

\subsubsection{Datasets}
To evaluate the effectiveness of our method, we conduct extensive experiments on two benchmarks, i.e., Flickr30K\cite{young2014image} dataset 
and MSCOCO \cite{lin2014microsoft} dataset, which are commonly employed in ITR. 
Specifically, Flickr30K and MSCOCO include 31,783 images and 123,287 images, respectively, and each image is annotated with five matched descriptions.
Following the protocol of \cite{lee2018stacked,faghri2018vse++}, Flickr30K is split into 29,000 training images, 1,000 validation images, and 1,000 testing images.
MSCOCO is split into 113,287 training images, 1,000 validation images, and 5,000 testing images.
It should be noted that the result of MSCOCO are reported in 5K and 1K, where the 1K results are averaged over the five 1K data folds. 
\subsubsection{Evaluation Metrics}
Following common practice in information retrieval \cite{faghri2018vse++,liu2021hit,han2021text}, we evaluate the retrieval performance by popular R@K (K=1,5,10) and R@Sum 
in terms of caption retrieval (image query) and image retrieval (caption query). Specifically, R@K indicates the proportion of queries whose ground-truth
is ranked within the top, and R@Sum is the summation of all six R@K, which illustrates the overall retrieval performance. We select the model with the highest R@Sum
on the Validation set for testing.

\subsection{Implementation Details}
For Query/Key Image Encoder, we employ the Faster-RCNN \cite{ren2016faster} with ResNet-101 to extract $K=36$ region proposals, which is pre-trained by Anderson \textit{et al.} 
\cite{anderson2018bottom} on Visual Genomes dataset \cite{krishna2017visual}, and the dimensionality of the region-level features is 2048 ($d^I=2048$).
In CGE module, we additionally leverage the CLIP (ViT/B-32) \cite{radford2021learning} model to 
extract the visual semantic embedding as the global representation ($d^I_c=512$). Note that the CLIP (ViT/B-32) model is not finetuned owing to the limited computational resources. 
For Query/Key Text Encoder, we adopt the BERT-base \cite{kenton2019bert} model to extract 768-dimensional word-level features ($d^T=768$).
The dimensionality of the joint embedding space is 1024 ($d^J=1024$). 
The momentum coefficient is set to 0.999 ($m=0.999$), and the sizes of dynamic queues for the MSCOCO and Flickr30K datasets are set to 4096 and 2048, respectively.
For the training stage, we empirically set $\gamma=90$ and $\epsilon=0.5$ in Eqs. \eqref{eq15}-\eqref{eq17}. 
Our method is trained by AdamW \cite{loshchilov2018decoupled} optimizer with weight decay factor $10e^{-4}$.
For MSCOCO, we train the model for 25 epochs. The initial learning rate is $5e^{-4}$, decaying by 10 for the last 10 epochs.
For Flickr30K, we train the model for 20 epochs. The initial learning rate is $5e^{-4}$, decaying by 10 for the last 10 epochs.
We employ PyTorch to implement our method, and our experiments are conducted on Intel(R) Xeon(R) CPU E5-2699 v4 @ 2.20GHz, single Tesla V100 GPU. 

\subsection{Comparisons with State-of-the-Art Methods}
We compare the proposed method with the state-of-the-art competing methods, which are roughly divided into global-level matching methods and local-level matching methods.
In short, the local-level matching methods need to calculate the similarity between the region-level and word-level representations.
However, global-level matching methods only require to calculate the similarity between two holistic representations. 
Specifically, the local-level matching methods include SCAN \cite{lee2018stacked}, CAMP \cite{wang2019camp}, IMRAM \cite{chen2020imram}, GSMN \cite{liu2020graph},
SHAN \cite{jistep}, DIME \cite{qu2021dynamic}, DSRAN \cite{wen2020learning},
SGRAF \cite{diao2021similarity}, UARDA \cite{zhang2022unified}, and NAAF \cite{zhang2022negative}.
The global-level matching methods include VSE++ \cite{faghri2018vse++}, VSRN \cite{li2019visual}, CVSE \cite{wang2020consensus}, MEMBER \cite{li2022image},
VSE$\infty$ \cite{chen2021learning}, and VSRN++ \cite{li2022image}.
Note that the proposed method belongs to the global-level matching method.

Since the feature aggregation in our method is developed according to the GPO in VSE$\infty$ \cite{chen2021learning}, we regard it as our baseline
model and report its replicated results via employing their open-sourced code with no change, which is marked with $\ddagger $ symbol in Tables \ref{table1}-\ref{table3}.
Additionally, several ensemble methods improve the performance via averaging the similarity from different models, which are marked with $\star $ symbol in Tables \ref{table1}-\ref{table3}.
\subsubsection{Results on MSCOCO dataset}
As shown in Tables \ref{table1} and \ref{table2}, we quantitatively report our results as well as all competing approaches on the MSCOCO 5-fold 1K Testing set and 
5K Testing set, respectively. Furthermore, the setting information of the models is also listed, such as image backbones and text backbones.
Note that the retrieval performance over relatively large dataset is also important, i.e., MSCOCO 5K Testing set, whereas it is often disregarded and not investigated
in previous methods, such as SHAN \cite{jistep}, CVSE \cite{wang2020consensus}, and GSMN \cite{liu2020graph}.  
From Tables \ref{table1}-\ref{table2}, we have several observations and conclusions:
\begin{itemize}
  \item {For the MSCOCO 5-fold 1K Testing set, the results are computed by averaging over 5 folds of 1K test samples. 
  Our USER (SGE) noticeably surpasses all the competing approaches on most evaluation metrics.
  Specifically, compared with NAAF \cite{zhang2022negative},
  the existing best method of local-level matching, USER (SGE) significantly improves it by 2.3\% and 2.0\% on R@1 at caption retrieval and image retrieval, respectively.
  Compared with VSRN++, the existing best method of global-level matching, USER (SGE) obtains 6.9\% improvement on R@Sum, where the R@1 obtains 4.9\% and 2.0\% improvement
  at caption retrieval and image retrieval.}
  \item {For the MSCOCO 5K Testing set, our USER (SGE) also achieves the best performance and significantly outperforms the other methods. 
  Concretely, compared with the best competitor NAAF \cite{zhang2022negative}, our USER obtains 14.6\% improvement on R@Sum, where the R@k (k=1,5,10) 
  gains (4.8\%, 2.2\%, 1.5\%) and (2.3\%, 2.5\%, 1.3\%) at caption retrieval and image retrieval, respectively. More importantly, we observe that our USER (SGE) obtains more
  significant improvement on the MSCOCO 5K Testing set than the MSCOCO 5-fold 1K Testing set, which exactly reflects the effect of increasing the scale of 
  negative pairs over larger testing set.} 
  \item {We observe considerable amount of improvements in USER (CGE) on both testing settings, especially in the MSCOCO 5K Testing set,
  where R@Sum obtains 10.5\% improvement over USER (SGE) and 25.1\% over NAAF. These results indicate that we obtain more discriminative global semantic 
  information through the CGE module, which can also be attributed to the large amount of pre-trained data WIT \cite{radford2021learning} employed by CLIP.}
\end{itemize}
\subsubsection{Results on Flickr30K dataset} In Table \ref{table3}, we show the results of the Flickr30K 1K Testing set, where our USER still performs the best.
Specifically, the results of USER (SGE) achieve the highest performance of 82.7\% and 63.1\% on R@1 at caption retrieval and image retrieval. 
With the help of the pre-trained model, our USER (CGE) further boosts the retrieval performance, in which the R@1 obtains 86.3\% and 69.5\% at 
caption retrieval and image retrieval.

\begin{table*}[!htb]
  \centering
  \caption{Quantitative Evaluation Results of Caption Retrieval and Image Retrieval on MSCOCO 5-fold 1K Testing Set. The Best and Second-Best Results Are Hightlighted in Bold and 
  Underlined. }
  \renewcommand\arraystretch{1.2}
  \begin{tabular}{lcccccccccccc}
      \toprule
      \multirow{2}{*}{\makecell{Learning\\ Paradigm}} & \multirow{2}{*}{Methods} & \multirow{2}{*}{Image Backbone} & \multirow{2}{*}{Text Backbone} & \multicolumn{3}{c}{Caption retrieval} & \multicolumn{3}{c}{Image retrieval} & \multirow{2}{*}{R@Sum}\\
      &&&& R@1 & R@5& R@10& R@1 & R@5& R@10\\
      \midrule
      \multirow{10}{*}{\makecell{Local-level\\matching}}&SCAN$_{\rm ECCV'18}^\star $ \cite{lee2018stacked}  & Faster-RCNN & BiGRU & 72.7& 94.8& 98.4& 58.8& 88.4& 94.8& 507.9 \\
      &CAMP$_{\rm ICCV'19}$ \cite{wang2019camp} & Faster-RCNN & BiGRU & 72.3 &94.8& 98.3& 58.5& 87.9& 95.0& 506.8 \\
      &IMRAM$_{\rm CVPR'20}^\star $ \cite{chen2020imram} & Faster-RCNN & BiGRU & 76.7& 95.6& 98.5& 61.7& 89.1& 95.0& 516.6\\
      &GSMN$_{\rm CVPR'20}^\star$ \cite{liu2020graph} & Faster-RCNN & BiGRU & 78.4& 96.4& 98.6& 63.3& 90.1& 95.7& 522.5 \\
      &SHAN$_{\rm IJCAI'21}^\star$\cite{jistep}  & Faster-RCNN & BiGRU & 76.8 & 96.3 & 98.7 &62.6 &89.6 &95.8& 519.5 \\
      &DIME$_{\rm SIGIR'21}$ \cite{qu2021dynamic} & Faster-RCNN & BERT & 77.9 & 95.9 & 98.3 & 63.0& 90.5& 96.2 &521.8 \\
      &DSRAN$_{\rm TCSVT'21}$ \cite{wen2020learning} & Faster-RCNN & BERT & 77.1 &95.3 &98.1 &62.9 &89.9& 95.3& 518.6 \\
      &SGRAF$_{\rm AAAI'21}^\star$ \cite{diao2021similarity} & Faster-RCNN & BiGRU & 79.6 & 96.2 &98.5& 63.2& 90.7 &96.1& 524.3 \\
      &UARDA$_{\rm TMM'22}$ \cite{zhang2022unified}  & Faster-RCNN & BiGRU & 78.6& 96.5& \underline{98.9} & 63.9& 90.7& \underline{96.2}& 524.8 \\
      &NAAF$_{\rm CVPR'22}^\star$ \cite{zhang2022negative}  & Faster-RCNN & BiGRU & 80.5 & 96.5& 98.8 &64.1 &90.7 &\textbf{96.5}& 527.2 \\
      \midrule
      \multirow{8}{*}{\makecell{Global-level\\matching}} &VSE++$_{\rm BMVC'18}$ \cite{faghri2018vse++} &ResNet-152 & GRU & 64.6& 90.0 &95.7& 52.0& 84.3 &92.0& 478.8 \\
      &VSRN$_{\rm ICCV'19}$\cite{li2019visual} & Faster-RCNN & BiGRU & 76.2& 94.8& 98.2& 62.8& 89.7& 95.1& 516.8 \\
      &CVSE$_{\rm ECCV'20}$ \cite{wang2020consensus}  & Faster-RCNN & BiGRU & 74.8& 95.1& 98.3& 59.9& 89.4& 95.2& 512.7 \\
      &MEMBER$_{\rm TIP'21}^\star$\cite{li2021memorize} & Faster-RCNN & BERT & 78.5& \textbf{96.8} & 98.5 & 63.7 & 90.7&95.6 &523.8 \\
      &VSE$\infty$$_{\rm CVPR'21}^\ddagger $ \cite{chen2021learning}   & Faster-RCNN & BERT & 78.6 & 96.2 & 98.7 & 62.9 &90.8 & 96.1 & 523.3 \\
      &VSRN++$_{\rm TPAMI'22}$ \cite{li2022image} & Faster-RCNN & BERT & 77.9 & 96.0 &98.5& 64.1& \underline{91.0} &96.1 &523.6 \\
      &USER (SGE) & Faster-RCNN & BERT & \underline{82.8} &\textbf{96.8}& 98.8 & \underline{66.1}& 90.6 &95.6& \underline{530.5} \\
      &USER (CGE) & Faster-RCNN\&CLIP & BERT & \textbf{83.7} &\underline{96.7}& \textbf{99.0} & \textbf{67.8}& \textbf{91.2} &95.8& \textbf{534.2} \\
      \bottomrule
  \end{tabular}
  \label{table1}
\end{table*}

\begin{table*}[!htb]
  \centering
  \caption{Quantitative Evaluation Results of Caption Retrieval and Image Retrieval on MSCOCO 5K Testing Set. The Best and Second-Best Results Are Hightlighted in Bold and 
  Underlined. The Symbol ``-" Indicates no Results.}
  \renewcommand\arraystretch{1.2}
  \begin{tabular}{lcccccccccccc}
      \toprule
      \multirow{2}{*}{\makecell{Learning\\Paradigm}} & \multirow{2}{*}{Methods} & \multirow{2}{*}{Image Backbone} & \multirow{2}{*}{Text Backbone} & \multicolumn{3}{c}{Caption retrieval} & \multicolumn{3}{c}{Image retrieval} & \multirow{2}{*}{R@Sum}\\
      &&&& R@1 & R@5& R@10& R@1 & R@5& R@10\\
      \midrule
      \multirow{8}{*}{\makecell{Local-level\\matching}}&SCAN$_{\rm ECCV'18}^\star$ \cite{lee2018stacked} & Faster-RCNN & BiGRU & 50.4 & 82.2 &90.0 &38.6& 69.3& 80.4& 410.9 \\
      &CAMP$_{\rm ICCV'19}$ \cite{wang2019camp} & Faster-RCNN & BiGRU & 50.1 & 82.1& 89.7& 39.0& 68.9& 80.2& 410.0\\
      &IMRAM$_{\rm CVPR'20}^\star$ \cite{chen2020imram} & Faster-RCNN & BiGRU & 53.7& 83.2& 91.0& 39.7& 69.1& 79.8& 416.5\\
      &DIME$_{\rm SIGIR'21}$ \cite{qu2021dynamic}  & Faster-RCNN & BERT & 56.1& 83.2& 91.1& 40.2& 70.7& 81.4& 422.7 \\
      &DSRAN$_{\rm TCSVT'21}$ \cite{wen2020learning}  & Faster-RCNN & BERT & 53.7& 82.1& 89.9& 40.3& 70.9& 81.3& 418.2 \\
      &SGRAF$_{\rm AAAI'21}^\star$ \cite{diao2021similarity} & Faster-RCNN & BiGRU & 57.8& -- & 91.6 & 41.9&--& 81.3& -- \\
      &UARDA$_{\rm TMM'22}$ \cite{zhang2022unified} & Faster-RCNN & BiGRU & 56.2& 83.8& 91.3& 40.6& 69.5& 80.9& 422.3 \\
      &NAAF$_{\rm CVPR'22}^\star$ \cite{zhang2022negative} & Faster-RCNN & GloVe & 58.9& 85.2& 92.0& 42.5& 70.9& 81.4& 430.9 \\
      \midrule
      \multirow{7}{*}{\makecell{Global-level\\matching}}&VSE++$_{\rm BMVC'18}$ \cite{faghri2018vse++} &ResNet-152 & GRU & 41.3& 71.1& 81.2&30.3&59.4&72.4& 409.8\\
      &VSRN$_{\rm ICCV'19}$\cite{li2019visual} & Faster-RCNN & BiGRU & 53.0& 81.1& 89.4& 40.5& 70.6& 81.1& 415.7 \\
      &VSE$\infty$$_{\rm CVPR'21}^\ddagger $ \cite{chen2021learning}   & Faster-RCNN & BERT & 57.4& 84.6& 91.8& 41.3& 71.9& 82.6& 429.6 \\
      &MEMBER$_{\rm TIP'21}^\star$ \cite{li2021memorize} & Faster-RCNN & BERT& 54.5 & 82.3 & 90.1 & 40.9 & 71.0 & 81.8 & 420.6 \\
      &VSRN++$_{\rm TPAMI'22}$ \cite{li2022image} & Faster-RCNN & BERT & 54.7 & 82.9 &90.9 &42.0 &72.2 &82.7 &425.4 \\
      &USER (SGE) & Faster-RCNN & BERT & \underline{63.7} &\underline{87.4}& \underline{93.5} &\underline{44.8} &\underline{73.4}& \underline{82.7}& \underline{445.5} \\
      &USER (CGE) & Faster-RCNN\&CLIP & BERT & \textbf{67.6} &\textbf{88.4}& \textbf{93.5} & \textbf{47.7}& \textbf{75.1}& \textbf{83.7}& \textbf{456.0} \\
      \bottomrule
  \end{tabular}		
  \label{table2}
\end{table*}

\begin{table*}[!htb]
  \centering
  \caption{Quantitative Evaluation Results of Caption Retrieval and Image Retrieval on Flickr30K 1K Testing Set. The Best and Second-Best Results Are Hightlighted in Bold and 
  Underlined.}
  \renewcommand\arraystretch{1.2}
  \begin{tabular}{lcccccccccccc}
      \toprule
      \multirow{2}{*}{\makecell{Learning\\Paradigm}} & \multirow{2}{*}{Methods}& \multirow{2}{*}{Image Backbone} & \multirow{2}{*}{Text Backbone} & \multicolumn{3}{c}{Caption retrieval} & \multicolumn{3}{c}{Image retrieval} & \multirow{2}{*}{R@Sum}\\
      &&&& R@1 & R@5& R@10& R@1 & R@5& R@10\\
      \midrule
      \multirow{9}{*}{\makecell{Local-level\\matching}}&SCAN$_{\rm ECCV'18}^\star$ \cite{lee2018stacked} & Faster-RCNN & BiGRU & 67.4& 90.3& 95.8& 48.6& 77.7& 85.2& 465.0 \\ 
      &CAMP$_{\rm ICCV'19}$ \cite{wang2019camp} & Faster-RCNN & BiGRU & 68.1& 89.7& 95.2& 51.5& 77.1& 85.3& 466.8\\
      &IMRAM$_{\rm CVPR'20}^\star$ \cite{chen2020imram} & Faster-RCNN & BiGRU & 74.1& 93.0& 96.6& 53.9& 79.4& 87.2& 484.2\\
      &GSMN$_{\rm CVPR'20}^\star$ \cite{liu2020graph} & Faster-RCNN & BiGRU & 76.4& 94.3& 97.3& 57.4& 82.3& 89.0& 496.8 \\
      &SHAN$_{\rm IJCAI'21}^\star$ \cite{jistep} & Faster-RCNN & BiGRU & 74.6 &93.5& 96.9& 55.3& 81.3& 88.4& 490.0 \\
      &DIME$_{\rm SIGIR'21}$ \cite{qu2021dynamic} & Faster-RCNN & BERT & 77.4& 95.0& 97.4& 60.1& 85.5& 91.8& 507.2 \\
      
      &DSRAN$_{\rm TCSVT'21}$ \cite{wen2020learning} & Faster-RCNN & BERT & 75.3& 94.4& 97.6& 57.3& 84.8& 90.9& 500.3 \\
      &SGRAF$_{\rm AAAI'21}^\star$ \cite{diao2021similarity} & Faster-RCNN & BiGRU & 77.8& 94.1& 97.4& 58.5& 83.0& 88.8& 499.6 \\
      &UARDA$_{\rm TMM'22}$ \cite{zhang2022unified} & Faster-RCNN & BiGRU & 77.8& 95.0& 97.6& 57.8& 82.9& 89.2& 500.3 \\
      &NAAF$_{\rm CVPR'22}^\star$\cite{zhang2022negative} & Faster-RCNN & GloVe & 81.9& 96.1& \underline{98.3}& 61.0& 85.3& 90.6& 513.2 \\
      \midrule
      \multirow{8}{*}{\makecell{Global-level\\matching}}&VSE++$_{\rm BMVC'18}$ \cite{faghri2018vse++} &ResNet-152 & GRU & 52.9 & 80.5 & 87.2& 39.6& 70.1& 79.5& 409.8 \\
      &VSRN$_{\rm ICCV'19}$\cite{li2019visual}& Faster-RCNN & BiGRU & 71.3 & 90.6& 96.0& 54.7 &81.8& 88.2& 482.6 \\
      &CVSE$_{\rm ECCV'20}$\cite{wang2020consensus}& Faster-RCNN & BiGRU & 73.5& 92.1& 95.8& 52.9& 80.4& 87.8& 482.5 \\
      &MEMBER$_{\rm TIP'21}^\star$ \cite{li2021memorize}& Faster-RCNN & BERT & 77.5 &94.7&97.3&59.5&84.8&91.0&504.8\\
      &VSE$\infty$$_{\rm CVPR'21}^\ddagger $ \cite{chen2021learning}& Faster-RCNN & BERT & 78.0& 95.1& 97.4&	58.7& 84.1& 90.4&	503.7       \\
      &VSRN++$_{\rm TPAMI'22}$ \cite{li2022image}& Faster-RCNN & BERT & 79.2& 94.6& 97.5& 60.6& 85.6& 91.4& 508.9 \\
      &USER (SGE) & Faster-RCNN & BERT & \underline{82.7}&	\underline{97.0}&	\underline{98.3}&	\underline{63.1}&	\underline{86.7}&	\underline{92.1}&	\underline{519.9}  \\
      &USER (CGE) & Faster-RCNN\&CLIP &BERT& \textbf{86.3}& \textbf{97.6}& \textbf{99.4}&	\textbf{69.5}& \textbf{91.0}& \textbf{94.4}&\textbf{538.1} \\
      \bottomrule
  \end{tabular}		
  \label{table3}
\end{table*}

\subsection{Ablation Studies}
To verify the effectiveness of each component in our proposed USER, we systematically perform extensive ablation studies on both the MSCOCO and 
Flickr30K datasets, including the configurations of our USER method, the impact of mini-batch size and dynamic queue size, and the sensitivity analysis of parameters in the  
UTO.
\subsubsection{Configurations of USER framework} To begin with, we first validate the effectiveness of each component in our USER 
method by conducting experiments on the Flickr30K 1K Testing set, as shown in Table \ref{table4}. For fair comparison, we employ the bidirectional triplet ranking \cite{faghri2018vse++} 
and report the results under the same hardware environment in Method I. In Method II, we employ the UTO objective function with mini-batch based training strategy. 
Our proposed GSE, DVQ, and DTQ are separately incorporated to demonstrate their effectiveness in Methods III, IV, and V, respectively. 
Based on Method II, we employ the combination of DVQ and DTQ in Method VI. Finally, Method VII is our full USER model.

Specifically, comparing Method I with Method II based on R@1, adding UTO substantially brings about 1.3\% improvement for caption retrieval and 2.9\% improvement for image retrieval.
Based on this, comparing Method II with Method V, the GSE module leads to 2.7\% improvement on R@Sum. 
Moreover, DVQ and DTQ alone obtain additional performance improvement, and the combination of them obtain even more significant improvement, 
as shown in Methods III, IV, and VI. 
\subsubsection{Impact of mini-batch and dynamic queue sizes}
The proposed UTO provides a novel training paradigm, in which different mini-batch sizes and dynamic queue sizes may influence the retrieval performance. 
We systematically study the impact of mini-batch size and dynamic queue size, and the experiment results are conducted on both the MSCOCO 5K Testing set and Flickr30K 1K Testing set.

As shown in Table \ref{table4}, the first four rows illustrates the impact of dynamic queue size on retrieval performance. 
For the Flickr30K dataset, when the dynamic queue size is 2048, the R@Sum that evaluates overall retrieval performance reaches the maximum value of 519.9.
For the MSCOCO dataset, when the dynamic queue size is 4096, the R@Sum reaches the maximum value of 445.5. 
On this basis, the last four rows shows the impact of mini-batch size on retrieval performance. 
Unlike dynamic queue size, mini-batch size has more significant impact on retrieval performance. When the batch size is 128, both datasets achieve the best retrieval performance, 
and when it is too large or too small, the retrieval performance of our model is affected negatively.

\subsubsection{Sensitivity analysis of parameters in UTO}
\begin{figure}[!htb]
  \centering
  \includegraphics[width=0.99\columnwidth]{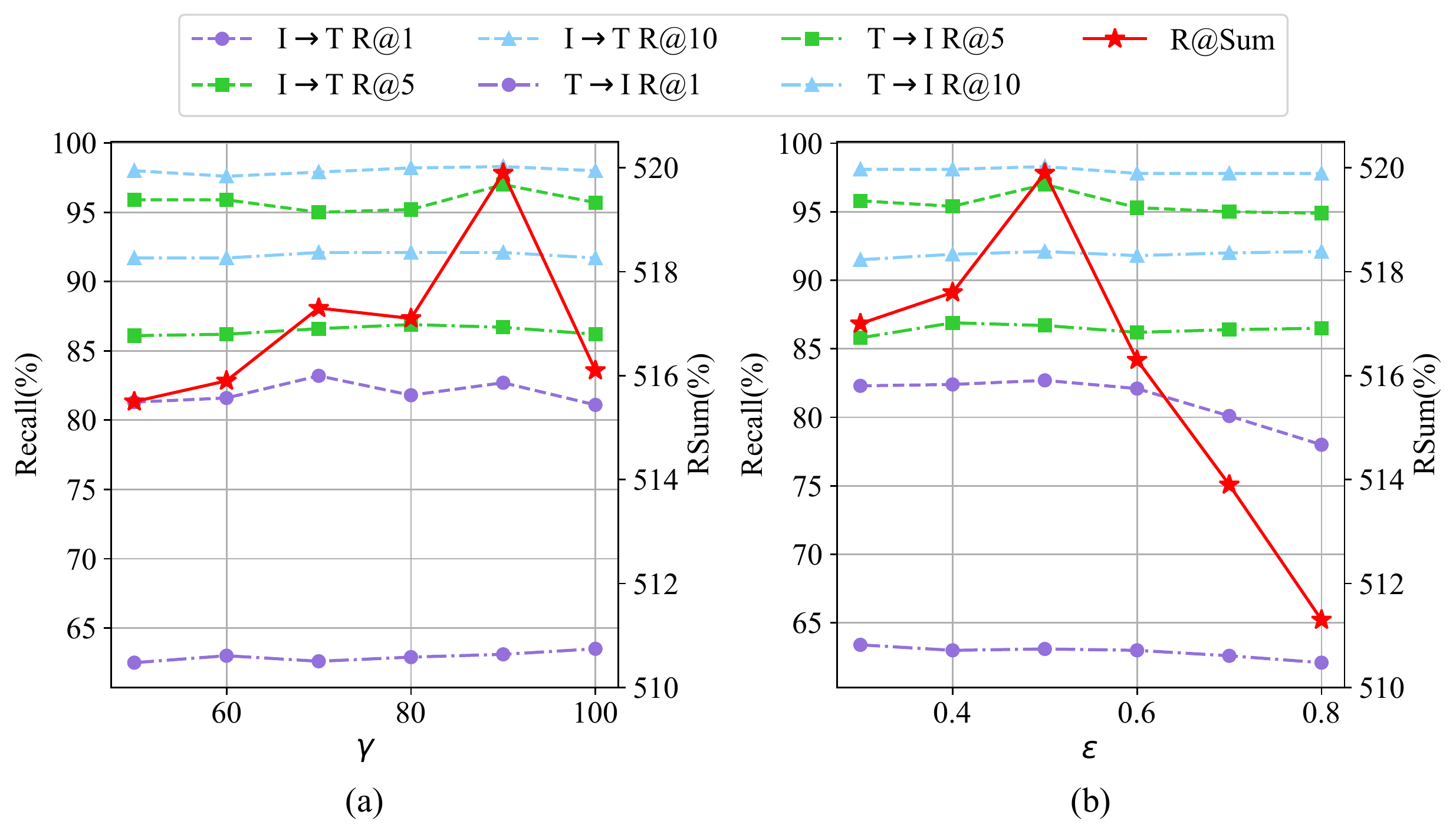} 
  \caption{Impacts of (a) $\gamma $ and (b) $\epsilon $ in $\mathcal{L}_{mini-hal} $, $\mathcal{L}_{DVQ-hal} $, and $\mathcal{L}_{DTQ-hal} $ on the Flickr30k dataset. 
          Note that R@k (k=1,5,10) refer to the left vertical coordinates while R@Sum refers to the right vertical coordinates.}
  \label{fig4}
\end{figure}

\begin{figure}[!htb]
  \centering
  \includegraphics[width=0.99\columnwidth]{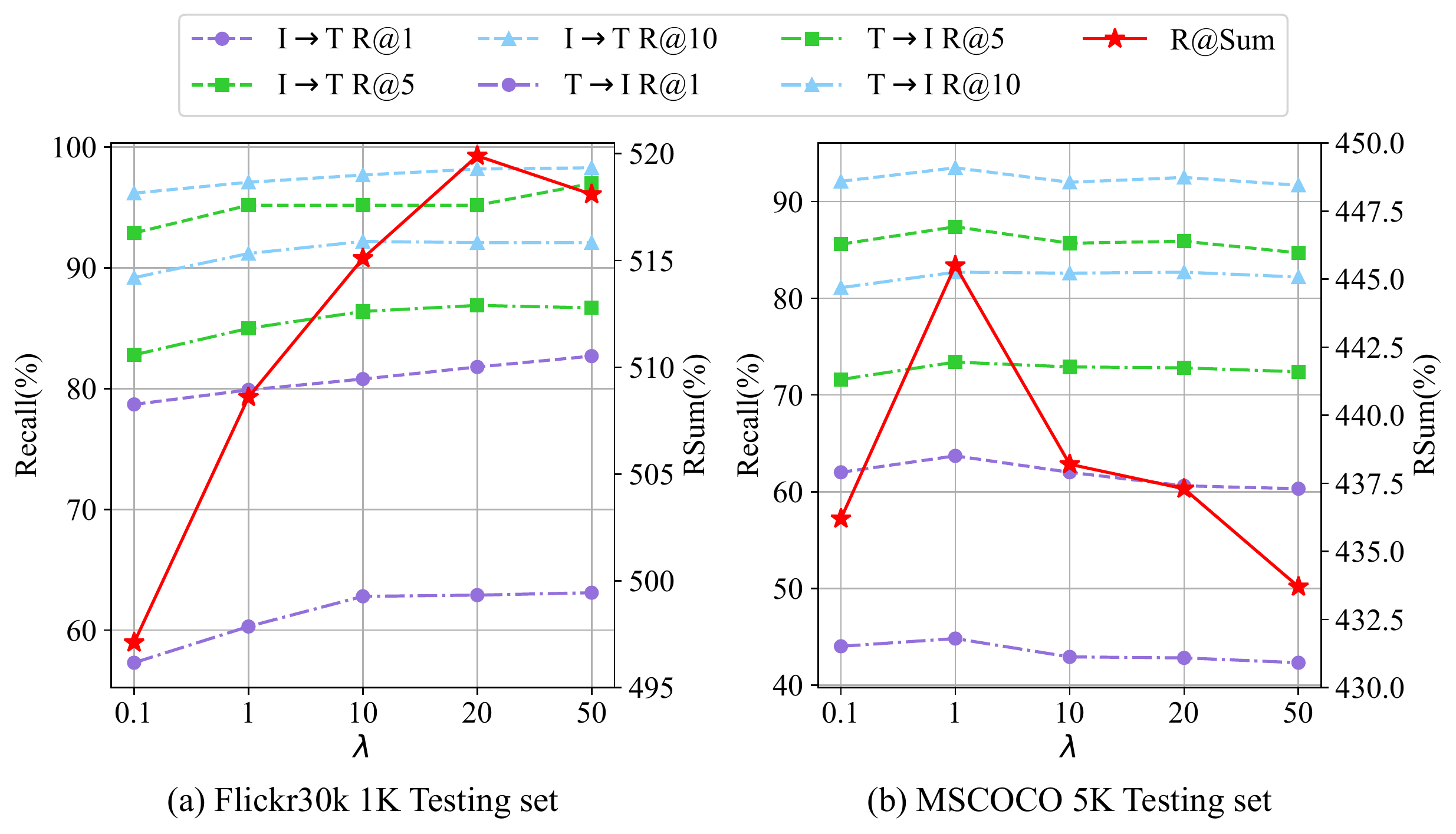} 
  \caption{Impacts of  $\lambda$ in $\mathcal{L} $ on the (a) Flickr30K and (b) MSCOCO datasets.
  Note that R@k (k=1,5,10) refer to the left vertical coordinates while R@Sum refers to the right vertical coordinates.}
  \label{fig5}
\end{figure}

As illustrated in Eqs. \eqref{eq15}-\eqref{eq19}, different types of negative sample pairs (i.e., mini-batch based and dynamic queue based) involve three hyperparameters, 
namely $\gamma$, $\epsilon$, and $\lambda$, respectively. $\gamma$ and $\epsilon$ constrain the loss weights of negative sample pairs, and $\lambda$ represents the 
different weights of $\mathcal{L}$. 
As shown in Fig. \ref{fig4} and Fig. \ref{fig5}, we experimentally observe that different hyperparameters affect the retrieval performance on both the MSCOCO dataset and Flickr30K dataset.

From Fig. \ref{fig4}(a), we could observe that the best $\gamma$ for R@Sum is 90, and the other metrics (R@1, R@5, R@10) also reach a relatively high level. Besides,
as the $\gamma$ increases, the model performance for R@Sum first increases and then decreases. The reason is that the large weights will affect the alignment of positive sample pairs. 
Similarly, the small weights will cause insufficient attention attended to negative sample pairs.
From Fig. \ref{fig4}(b), we could observe that the best $\epsilon$ for R@Sum is 0.5. Similar to $\gamma$, R@Sum also increases first and then decreases as $\epsilon$ increases.

$\mathcal{L} _{mini-hal}$ and $\mathcal{L} _{DQ-hal}$ are combined by $\lambda$ in Eq. \eqref{eq19}, where $\lambda$ balances the two objective functions. 
As shown in Fig. \ref{fig5}, we further experimentally select appropriate hyperparameters on the MSCOCO and Flickr30K datasets, respectively. 
Specifically, we select different hyperparameters $\lambda$ from the set $\boldsymbol{\Lambda} \in \{0.1,1,10,20,50\}$. From Fig. \ref{fig5}, the best $\lambda$ of R@Sum is 20 for the Flickr30K dataset, 
while 1 for the MSCOCO dataset. The reason for this phenomenon may be that the MSCOCO dataset is much larger than the Flickr30K dataset, and relatively speaking, more negative samples from 
the dynamic queue are needed for learning.

\setlength{\tabcolsep}{1pt}
\begin{table}[!htb]
  \centering
  \caption{Performance Comparison of Our USER with Different Main Components on Flickr30K 1K Testing Set.}
  \renewcommand\arraystretch{1.2}
  \begin{tabular}{ccccccccccccc}
      \toprule
      \multirow{2}{*}{Models}&\multirow{2}{*}{DVQ}&\multirow{2}{*}{DTQ} &\multirow{2}{*}{GSE} &\multirow{2}{*}{UTO} & \multicolumn{3}{c}{Caption retrieval} & \multicolumn{3}{c}{Image retrieval} & \multirow{2}{*}{R@Sum}\\
      &&&&& R@1 & R@5& R@10& R@1 & R@5& R@10\\
      \midrule
      I &&&&& 78.0& 95.1& 97.4&	58.7& 84.1& 90.4&	503.7 \\ 
      II &&&&\checkmark& 79.3&	95.3&	97.6&	61.6&	86.0&	91.5&	511.3 \\ 
      III & \checkmark &&&\checkmark&80.0& 97.1& 98.3&63.0& 86.4& 92.3&517.1 \\
      IV &&\checkmark&&\checkmark&81.0 &95.0& 98.1& 61.4& 86.0& 91.6&513.1 \\
      V &&&\checkmark&\checkmark& 80.6 &95.5& 97.6& 62.2& 86.4& 91.6& 514.0\\
      VI &\checkmark&\checkmark&&\checkmark& 82.0& 95.9& 98.0&63.0& 87.1& 92.2&518.2 \\
      VII &\checkmark&\checkmark&\checkmark&\checkmark& 82.7&	97.0&	98.3&	63.1&	86.7&	92.1&	519.9  \\
      \bottomrule
  \end{tabular}		
  \label{table4}
\end{table}

\setlength{\tabcolsep}{4pt}
\begin{table*}[t]
  \centering
  \caption{Performance Comparison of Different Dynamic Queue Sizes and Mini-batch Sizes on Flickr30K 1K and MSCOCO 5K Testing Sets.}
  \renewcommand\arraystretch{1.2}
  \begin{tabular}{c|c|cccccccccccccc}
      \toprule
      \multirow{3}{*}{Type} & \multirow{3}{*}{Size}&  \multicolumn{7}{c}{Flickr30K 1K Testing set} & \multicolumn{7}{c}{MSCOCO 5K Testing set} \\
      &&\multicolumn{3}{c}{Caption retrieval} & \multicolumn{3}{c}{Image retrieval} & \multirow{2}{*}{R@Sum} &\multicolumn{3}{c}{Caption retrieval} & \multicolumn{3}{c}{Image retrieval} & \multirow{2}{*}{R@Sum} \\
      && R@1 & R@5& R@10& R@1 & R@5& R@10 &&R@1 & R@5& R@10& R@1 & R@5& R@10\\
      \midrule
      \multirow{4}{*}{\makecell{Dynamic Queue\\ Size}}&512 & 82.0& 96.0& 98.3&62.4& 86.9& 91.8&517.4& 60.5& 85.7& 91.9&42.8& 72.5& 82.1&435.6\\ 
      &1024 & 80.9& 95.1& 97.7&62.5& 86.4& 92.1&514.7& 62.5& 86.4& 92.5&44.2& 72.8& 82.2&440.5 \\
      &2048 & 82.7&	97.0&	98.3&	63.1&	86.7&	92.0&519.9&63.2& 87.0& 93.1&44.9& 73.5& 82.4&444.0\\
      &4096 & 82.3& 95.4& 97.8 &63.1 &86.7& 91.9&517.2&63.7 &87.4& 93.5 &44.8 &73.4& 82.7& 445.5\\
      \midrule
      \multirow{4}{*}{\makecell{Mini-batch\\ Size}}&32 &75.9& 93.3& 97.7&58.8& 84.1& 90.9&500.7&59.9& 84.7& 91.6&43.5& 72.6& 82.2&434.5\\
      &64 & 80.0& 95.1& 97.7&61.4& 86.4& 92.1&512.6& 61.9& 87.2& 93.3&44.8& 73.3& 82.7&443.2\\
      &128 & 82.7&	97.0&	98.3&	63.1&	86.7&	92.1&	519.9&63.9 &87.0& 93.4 &44.9 &73.6& 82.6& 445.4\\
      &256 & 81.4& 94.9& 98.0&62.9& 86.5& 91.7&515.4&63.7 &87.4& 93.5 &44.8 &73.4& 82.7& 445.5\\
      \bottomrule
  \end{tabular}		
  \label{table5}
\end{table*}

\subsection{Analysis on Inference Efficiency}
In practice, inference efficiency is as essential as retrieval performance, which is disregarded in existing methods.
We thereby provide a comprehensive and detailed comparison of the inference efficiency of existing mainstream algorithms.
First, we divide inference time into encoding time and matching time, where the former
represents the time cost for calculating the embedding, and the latter represents the time cost for computing
the similarity. Six current representative comparison algorithms, whose codes are publicly available, are tested for their
inference time under the same environment.
Specifically, the SCAN \cite{lee2018stacked}, SHAN \cite{jistep}, SGRAF \cite{diao2021similarity}, 
and NAAF \cite{zhang2022negative} methods belong to the local-level matching method, and the VSRN \cite{li2019visual} and VSE$\infty$ \cite{chen2021learning} methods belong to the global-level matching method.
For the ensemble methods, we report the inference time on single model since the ensemble results are calculated by two offline
similarity matrices.

\begin{figure}[!htb]
  \centering
  \includegraphics[width=0.99\columnwidth]{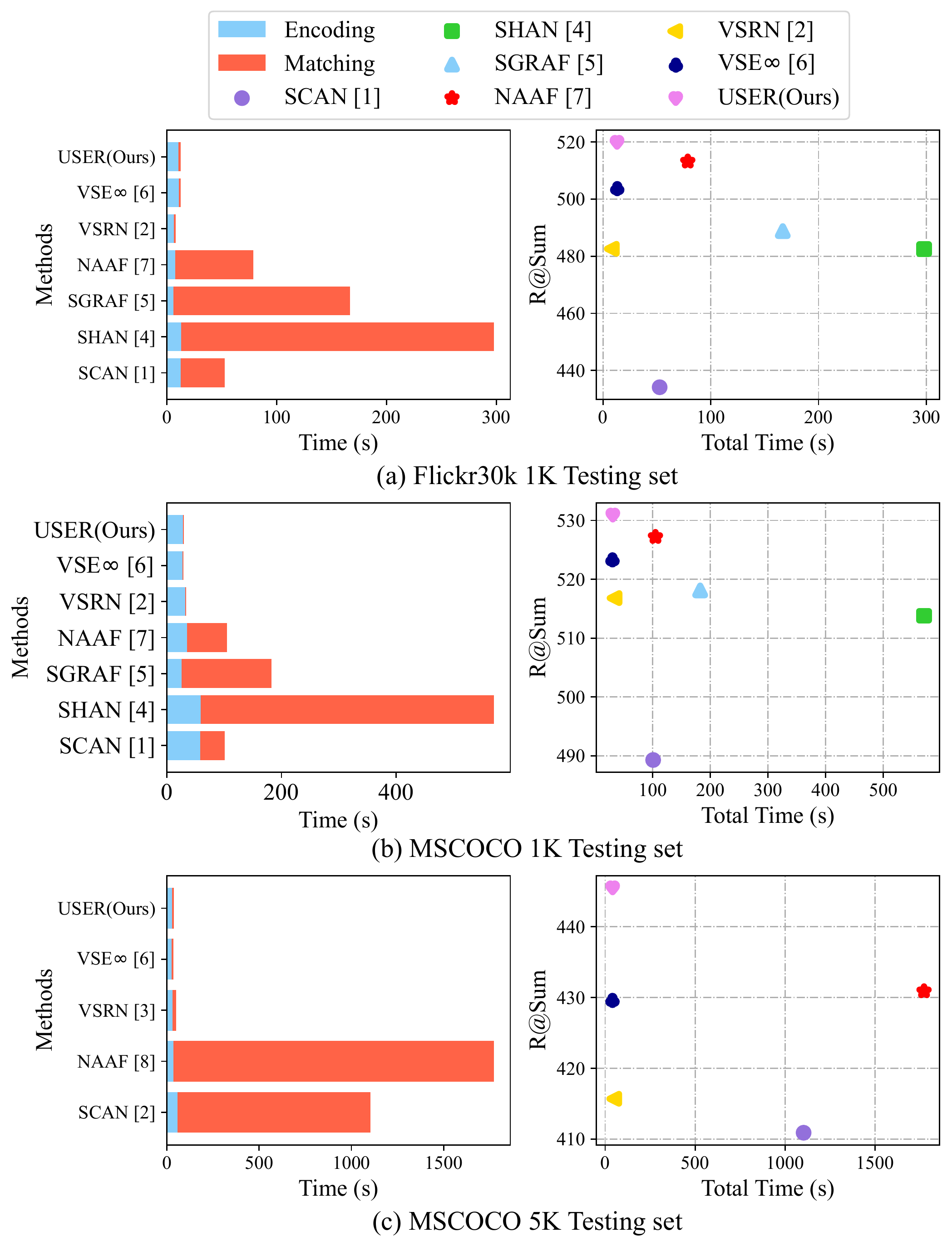} 
  \caption{The comparisons of the inference time with recent six methods on the (a) Flickr30K 1K Testing set, 
  (b) MSCOCO 5-fold 1K Testing set, and (c) MSCOCO 5K Testing set, respectively.}
  \label{fig6}
\end{figure}

As shown in Fig. \ref{fig6}, we report the inference time of our USER method with six competing methods on the Flickr30K 1K Testing set, 
MSCOCO 5-fold 1K Testing set, and MSCOCO 5K Testing set, respectively.
The three subfigures on the left illustrate the proportion of encoding time and matching time, and the comparison of different methods.
The three subfigures on the right illustrate the R@Sum-Time tradeoff of single model, where the closer to the upper left corner,
the better the overall performance. From the Fig. \ref{fig6},
several observations and conclusions are summarized as follows:
\begin{itemize}
  \item {For inference time, the global-level matching methods are much faster than local-level matching methods since the latter employs 
  complex local matching strategies such as the attention-weighted algorithms in SCAN \cite{lee2018stacked}, 
  the negative-aware attention in NAAF \cite{zhang2022negative}, while the former only needs to compute the
  similarity score by matrix multiplication.
  Specifically, the inference time of SHAN \cite{jistep}, SGRAF \cite{diao2021similarity}, and NAAF \cite{zhang2022negative} on the Flickr30K 1K Testing set 
  are 297.9s, 166.8s, and 78.6s, respectively. Nevertheless, the inference time of our USER is only 10.8s,
  which is the same order of magnitude as the other two global-level matching algorithms, i.e., VSRN and VSE$\infty$}.
  \item {Mathematically, compared with NAAF \cite{zhang2022negative}, the existing best method, 
  the inference time of our USER is 7.3 times faster and R@Sum is improved by 6.7\% on the Flickr30K 1K Testing set, 
  3.7 times and 3.3\% on the MSCOCO 1K Testing set, and 61.3 times and 14.6\% on the MSCOCO 5K Testing set.
  NAAF \cite{zhang2022negative} takes more time than our USER when the database is relatively large. Since the NAAF relies on densely matching among 
  all region-word pairwise similarities while ours just expands the matrices for dot product.}
  \item {From another perspective, the embeddings can be calculated offline and cached in advance. Therefore, the shorter matching time means higher retrieval efficiency.
  For local-level matching methods, it still consumes a lot of time on the matching stage since it requires tedious cross-modal interaction. 
  For example, the matching time of SHAN on the Flickr30K 1K Testing set is 284.8s, and that of NAAF on the MSCOCO 5K Testing set is 1736.3s.
  On the contrary, the matching time of our USER on the MSCOCO 1K Testing set is only about 0.05s, 
  and about 0.45s on the MSCOCO 5K Testing set. To sum up, our USER is efficient in practice, especially when the encoding stage can
  be done offline in advance.
  }
\end{itemize}

\section{Conclusion}
This paper has presented a novel Unified Semantic Enhancement Momentum Contrastive Learning (USER) method for Image-Text Retrieval.
Within the proposed USER method, two semantic enhancement modules are developed to enhance the visual representation by generating the discriminative global representation.
Particularly, with the vigorous self-attention \cite{vaswani2017attention} algorithm, the Self-Guided Enhancement (SGE) module is designed to learn the global representation by considering the relevance between each region and
the global information. Besides, we provide another simple but effective method to learn the global representation, which turns to the pre-trained CLIP \cite{radford2021learning} module, called CLIP-Guided Enhancement
(CGE) module.
Moreover, aims at enlarging the scale of negative pairs and learning more precise representation, we incorporate the training mechanism of Momentum Contrast with an improved Unified Training Objective (UTO).
Finally, systematical experiments on two benchmark datasets demonstrate that our USER method achieves a new state-of-the-art or competitive performance. 
More importantly, our method achieves outstanding retrieval efficiency, especially on a large-scale database for image or text queries. For instance, the inference time on the MSCOCO 5K Testing set is
more than 60 times faster than the recent SOTA methods.

In the future, we plan to expand our method in two aspects. 
On the one hand, more robust and discriminative object detectors and advanced contrast learning algorithm could improve the performance. On the other hand, 
it is illuminating to deploy our method on other retrieval tasks, such as video-text retrieval and content-based image retrieval.

\bibliographystyle{IEEEtran}

\bibliography{USER.bib}

\begin{IEEEbiography}[{\includegraphics[width=1in,height=1.25in,clip,keepaspectratio]{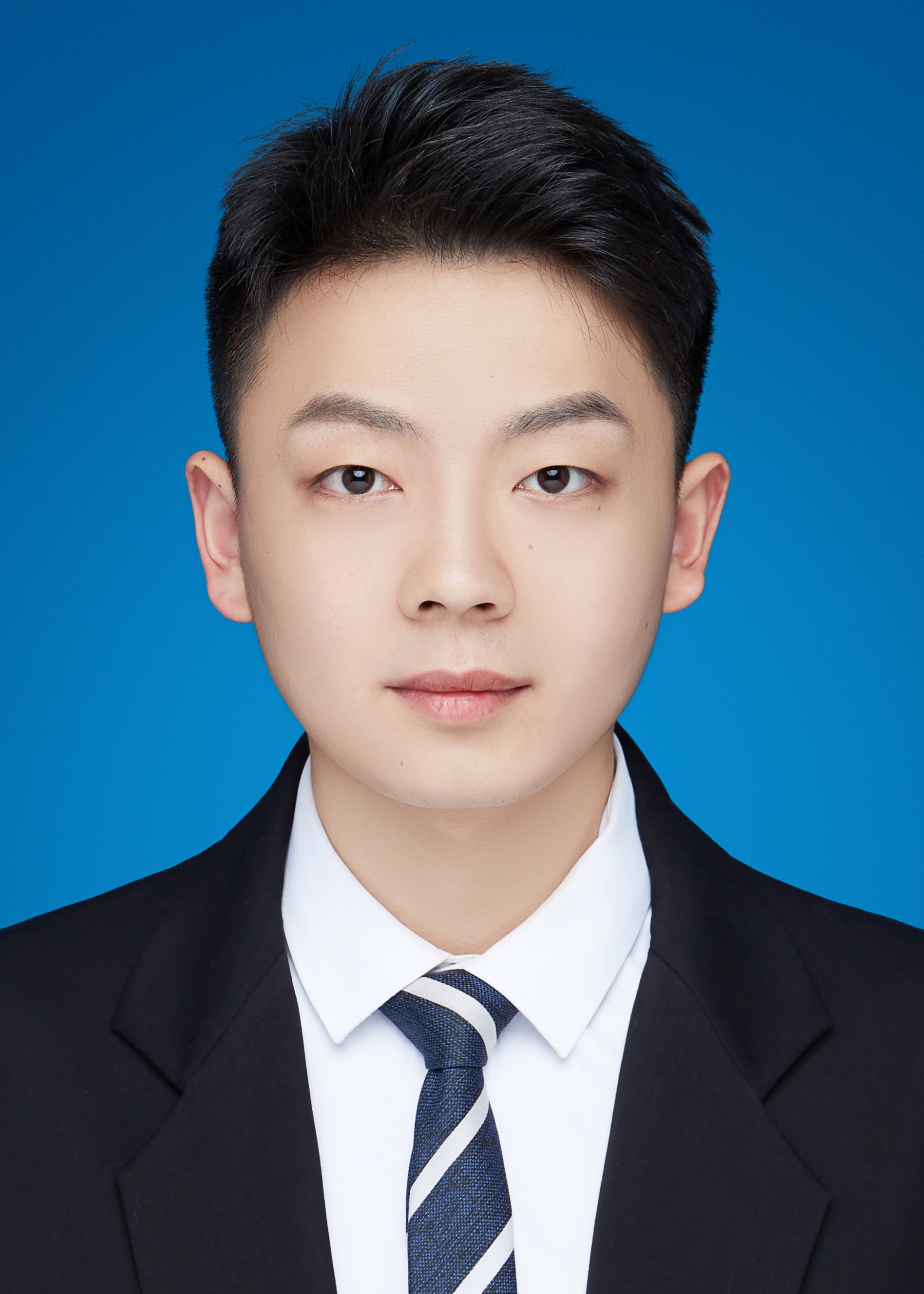}}]{Yan Zhang}
  received the M.S. degree in control engineering from Tianjin University,
  Tianjin, China, in 2021. He is currently pursuing
  a Ph.D. degree in the School of Electrical and Information
  Engineering, Tianjin University. His current research interests
  include cross-modal retrieval and computer vision.
\end{IEEEbiography}

\begin{IEEEbiography}[{\includegraphics[width=1in,height=1.25in,clip,keepaspectratio]{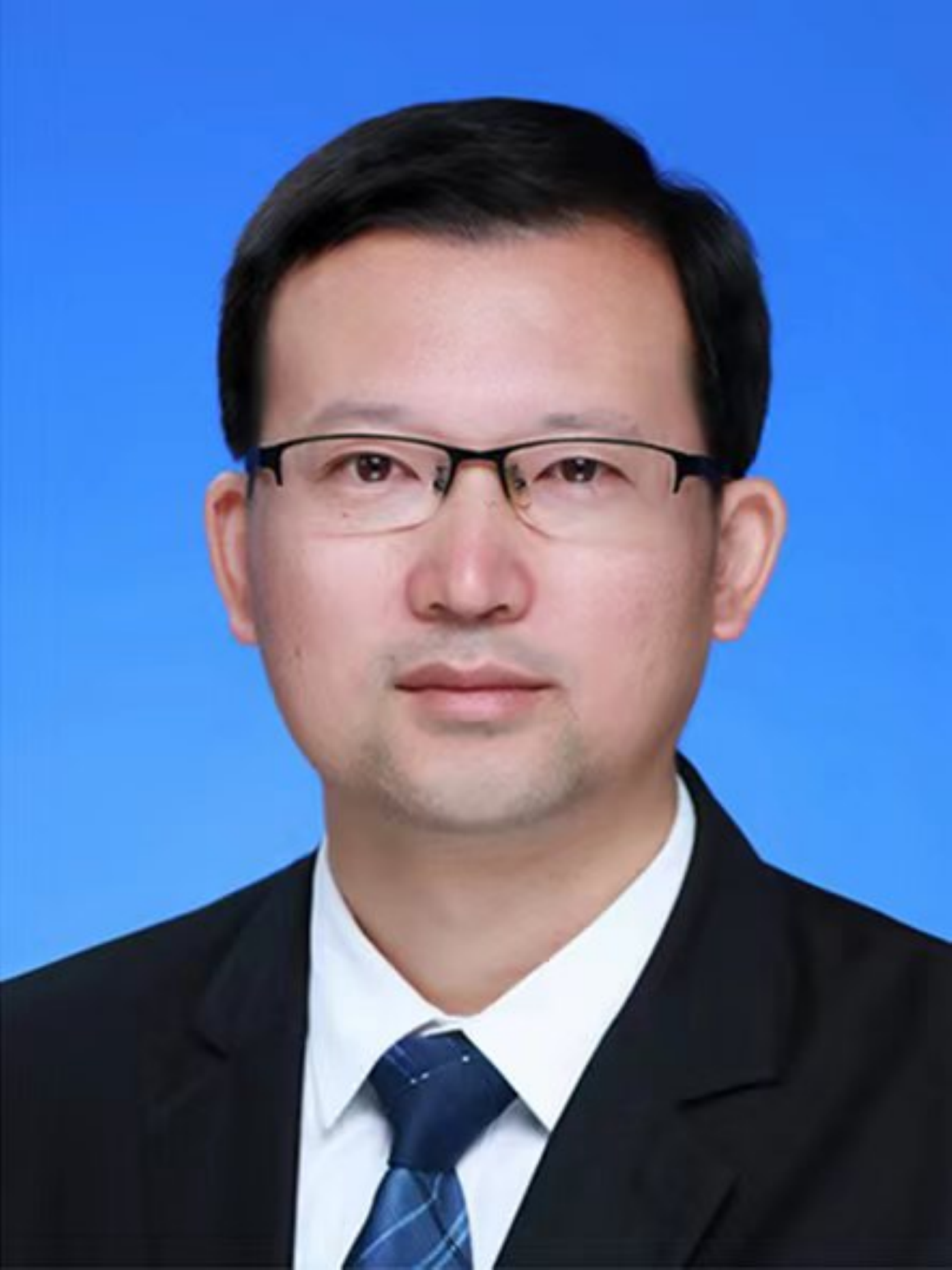}}]{Zhong Ji}
  received the Ph.D. degree in signal and information processing from Tianjin University, Tianjin, China, in 2008. He is currently a 
  Professor with the School of Electrical and Information Engineering, Tianjin University. He has authored over 100 technical articles 
  in refereed journals and proceedings. His current research interests include multimedia understanding, zero/few-shot learning, 
  and cross-modal analysis.
  \end{IEEEbiography}

\begin{IEEEbiography}[{\includegraphics[width=1in,height=1.25in,clip,keepaspectratio]{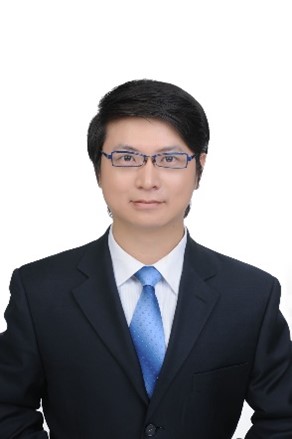}}]{Di Wang}
  received his Ph.D. degree from Tianjin University, China, in 2018. He is now a lecturer at School of Electrical and 
  Information Engineering, Tianjin University. His current research interests include conformal prediction and machine learning.
\end{IEEEbiography}

\begin{IEEEbiography}[{\includegraphics[width=1in,height=1.25in,clip,keepaspectratio]{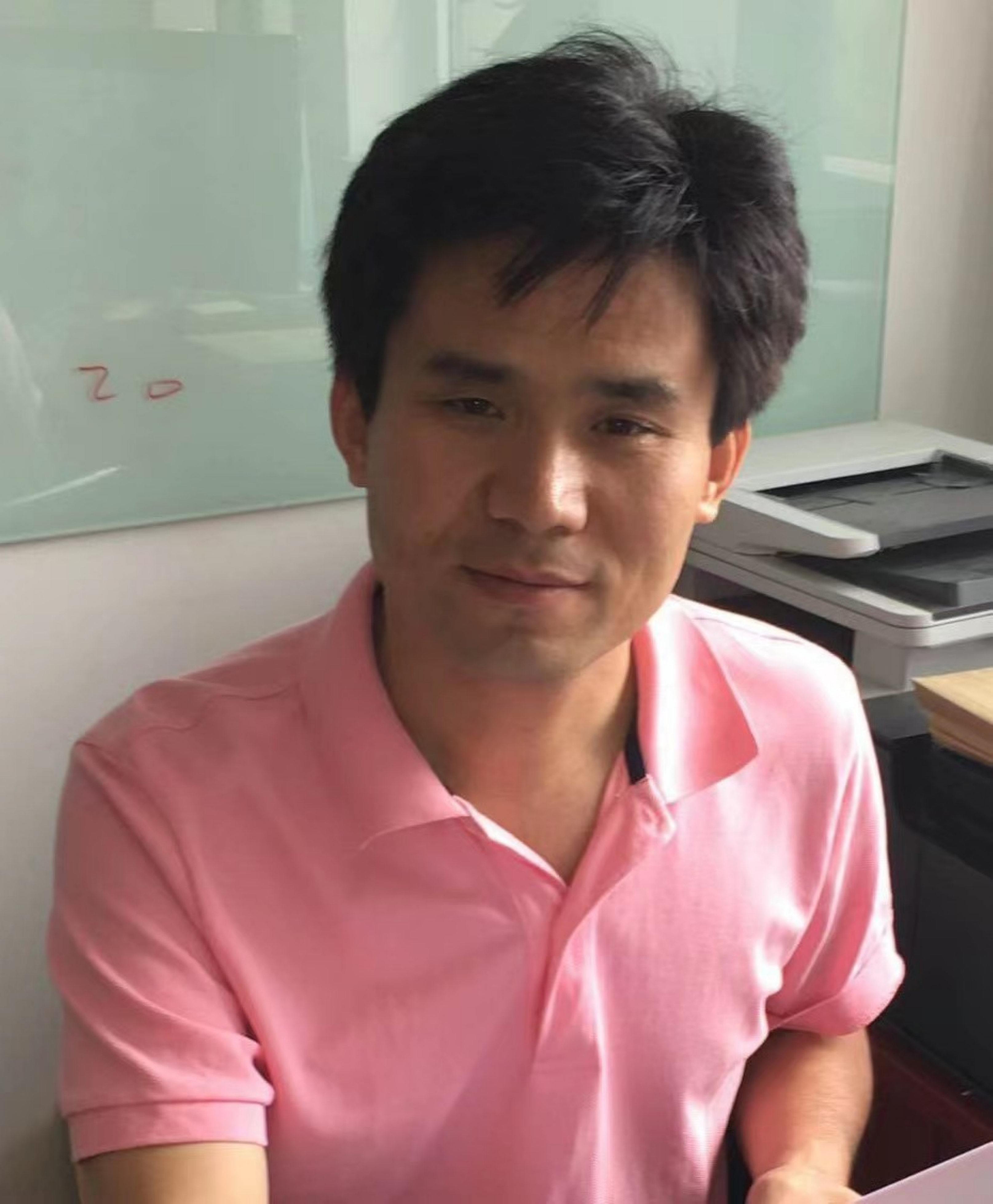}}]{Yanwei Pang}
  received the Ph.D. degree in electronic engineering from the University of Science and Technology of China, Hefei, China, in 2004.
  He is currently a Professor with School of Electrical and Information Engineering, Tianjin University, Tianjin, China. 
  He has authored over 150 technical articles in refereed journals and proceedings. His current research interests include object 
  detection and recognition, vision in bad weather, and computer vision.
  \end{IEEEbiography}

\begin{IEEEbiographynophoto}{Xuelong Li}
  (M'02-SM'07-F'12) is a full professor with School of Artificial Intelligence, OPtics and ElectroNics (iOPEN), Northwestern Polytechnical University, 
  Xi'an 710072, P.R. China.
  \end{IEEEbiographynophoto}

\end{document}